\documentclass{article}
\usepackage{ijcai26}
\usepackage{times}
\usepackage{soul}
\usepackage{url}
\usepackage{xurl}
\usepackage[hidelinks]{hyperref}
\usepackage[utf8]{inputenc}
\usepackage[small]{caption}
\usepackage{graphicx}
\usepackage{amsmath}
\usepackage{paralist}
\usepackage{amssymb}
\usepackage{amsmath}
\usepackage{upgreek}
\usepackage{multirow}
\usepackage{bm}
\usepackage{booktabs}
\usepackage{tabularx}
\usepackage{makecell}
\usepackage[ruled,vlined,linesnumbered]{algorithm2e}
\usepackage{algorithmic}
\usepackage{fontawesome5}
\usepackage{xspace}
\usepackage{xcolor}
\usepackage{subcaption}
\usepackage{placeins}
\usepackage{float}
\usepackage{nicefrac}
\usepackage{xfrac}
\usepackage{soul}
\usepackage[switch]{lineno}
\usepackage[normalem]{ulem} 

\newcommand{\sys}{\textsc{TestNav}\xspace}
\newcommand{\imagenetSymb}{\faCameraRetro\xspace}
\newcommand{\qqpSymb}{\faQuora\xspace}
\newcommand{\recodeSymb}{\faCode\xspace}
\newcommand{\mbppSymb}{\faLightbulb\xspace}

\title{\sys: Pareto-Guided Search for Compositional Robustness Testing}

\author{
Arooj Arif$^1$
\and
Tobias Hartung$^1$\and
Elena Botoeva$^{2}$\And
Alexandros Koliousis$^1$\\
\affiliations
$^1$Northeastern University London\\
$^2$University of Kent\\
\emails
\{arooj.arif, tobias.hartung\}@nulondon.ac.uk,
e.botoeva@kent.ac.uk,
alexandros.koliousis@nulondon.ac.uk
}

\begin{document}

\maketitle

\begin{abstract}
Deep learning models remain vulnerable
to real-world input perturbations,
especially when multiple corruptions
co-occur in the same input
(e.g., brightness shifts and motion
blur). Compositional testing reveals
these interaction effects but introduces
two challenges: combinatorial growth
of the perturbation space as
dimensions and severity levels increase,
and uneven diagnostic value---many
combinations yield unrealistically degraded
inputs with limited practical relevance.

We present \sys,\footnote{Code
and data are available in
an anonymised
\href{https://osf.io/kyb7g/overview?view_only=9a41411b99124f3d947712500e329da5}
{OSF repository}.}
a Pareto-guided robustness testing
framework for efficiently exploring
discrete, compositional perturbation
spaces when only a limited number
of perturbation configurations can
be evaluated.
\sys prioritises severe yet realistic
failures by formulating robustness
testing as bi-objective optimisation:
maximise performance degradation while
preserving input fidelity measured
by modality-specific metrics
(e.g., SSIM and KID for
vision; chrF and BERT-F1
for language and code).
It uses NSGA-II to approximate
the bi-objective Pareto front.
Across four benchmarks spanning vision,
natural language, and code generation,
\sys recovers Pareto fronts
up to $2.15\times$
faster than
search-based baselines, using  35.8\%--89.3\% of the discrete
perturbation space defined by four
perturbation dimensions with six levels
each.

\end{abstract}

\section{Introduction}
\label{sec:introduction}

Modern deep learning models achieve
strong task performance yet remain
vulnerable to semantics-preserving
input perturbations that induce model
failures~\cite{szegedy2014intriguing,goodfellow2015adversarial}.
Robustness testing typically evaluates
perturbations in isolation~\cite{imagenetc,mu2019mnistc}.
In practice, however, real-world
degradations often arise from interacting
factors (e.g., brightness shifts,
motion blur, and noise), which
rarely occur alone in deployment
settings such as autonomous
driving~\cite{musat2021multiweather,hao2024your}
and medical imaging~\cite{chuah2024towards}.

Compositional robustness testing---that is,
evaluating inputs under combinations of
multiple perturbations---reveals interaction
effects that remain undetected under
single-perturbation evaluation~\cite{mintun2021interaction,hendrycks2020augmix,testifai2026}.
However, systematically exploring
multi-perturbation configuration spaces
poses two key challenges:

\begin{inparaenum}[\it (i)]
\item
The perturbation space grows combinatorially:
with $n$ perturbation types and
$\ell$ discrete levels per type,
the number of configurations scales
as $\ell^n$.
For example, four perturbation types
with six levels each yield 
$6^4 = 1{,}296$ perturbation
configurations.
Exhaustive testing is computationally
expensive and scales poorly as
the number of perturbation types
or test inputs grows.

\item
Configurations are not equally informative:
some produce heavily degraded inputs
with low input fidelity---meaning the
perturbed input no longer resembles
the original---whereas others preserve
input fidelity yet induce model
failures.
Input fidelity does not degrade
monotonically with the number of
perturbations, rendering fixed-order
heuristics ineffective.
\end{inparaenum}


\textit{What makes a failure informative?}
The most informative failures are those where a perturbed input remains semantically faithful to the original yet still causes the model to fail, exposing genuine robustness weaknesses rather than expected sensitivity to obvious input corruption.
Finding such failures requires balancing \emph{two complementary objectives}: model performance degradation, which measures how much model performance drops under a given perturbation configuration, and input fidelity, which measures how closely perturbed inputs resemble the originals under appropriate similarity metrics.

For robustness testing, we are interested in the \emph{Pareto front} of perturbation configurations with respect to these two objectives. The Pareto front $\mathcal{P}^*$ consists of all \emph{Pareto-optimal} configurations---those where improving one objective necessarily worsens the other~\cite{miettinen1999}. Pareto-optimal configurations are the natural candidates for robustness testing: they maximise model performance degradation while preserving input fidelity, and are therefore most likely to reveal meaningful robustness failures.

%


We introduce \sys, a framework for compositional robustness testing that formalises this idea as a bi-objective optimization problem--balancing failure severity against input fidelity--and uses NSGA-II~\cite{deb2002fast}, a  multi-objective evolutionary algorithm, to efficiently approximate the Pareto front.
Our contributions are:
\begin{enumerate}[(1)]
\item We formalise compositional robustness testing as a budget-constrained bi-objective search problem over perturbation configurations, with model-performance degradation and input fidelity as competing objectives. We instantiate this formulation in \sys, an NSGA-II-based framework, which maintains a diverse set of non-dominated solutions along the degradation--fidelity trade-off.

\item We conduct an empirical study across four benchmarks---Tiny-ImageNet, QQP, HumanEval, and MBPP---spanning vision, language, and code. We exhaustively evaluate all $6^4$ configurations per dataset to construct ground-truth Pareto fronts, measuring fidelity via SSIM and KID (images) and chrF and BERTScore (language/code). Under identical budgets, \sys recovers these fronts 
up to $2.15\times$
faster than search-based baselines.
Tiny-ImageNet experiments further show that
neural-coverage metrics do not reliably
identify the Pareto-optimal failure
region.
  
\end{enumerate}

\section{Related Work}
\label{sec:related}

Deep learning robustness testing
evaluates model behaviour under
controlled input perturbations intended
to preserve task semantics while
shifting the input distribution.
Testing methods differ in what
they choose to vary:
\begin{inparaenum}[\it (i)]
\item
input-level methods generate, rank,
or assess individual test inputs
using activation coverage, uncertainty,
or distributional novelty; they
provide baseline search signals in
our study (\S\ref{sec:ilt});
\item
perturbation-level benchmarks evaluate
predefined corruption or transformation
families, typically one family at
a time; they provide the
perturbations used in our evaluation
(\S\ref{sec:plt});
\item
configuration-level methods evaluate
combinations of perturbations, inducing
a discrete space of perturbation
configurations; they motivate the
multi-perturbation space searched by
\sys (\S\ref{sec:clt}).
\end{inparaenum}

\sys builds on the configuration-level
view and uses multi-objective optimisation
as the algorithmic basis for
Pareto-guided search (\S\ref{sec:moo}).

\subsection{Input-level testing}
\label{sec:ilt}

Coverage-guided methods use activation
behaviour as a proxy for test
adequacy.
DeepXplore~\cite{deepxplore} formulates
DNN testing as an optimisation problem,
generating inputs that maximise neuron
activation coverage (NAC) across
multiple models.
It was among the first to
formulate test generation as guided
search, but its coverage criterion
is structural and does not measure
whether generated inputs remain close
to the originals.
DeepGauge~\cite{ma2018deepgauge} extends this
idea to multi-granularity activation
criteria, including strong neuron
activation coverage (SNAC), 
$k$-multisection neuron coverage (KMNC),
and top-$k$ neuron coverage (TKNC).
This family of methods guides testing
in input space rather than
perturbation-configuration space.

Input-prioritisation methods instead rank
candidate inputs by fault-revealing
signals.
DeepGini~\cite{deepgini} prioritises inputs
on which the model has low confidence,
using prediction uncertainty as a proxy
for fault-revealing potential.
Surprise Adequacy~\cite{kim2019surprise} ranks
inputs by how distributionally unusual
they are relative to training data,
using likelihood-based surprise adequacy
(LSA) or distance-based surprise adequacy
(DSA).
These methods help select individual
test inputs, but do not explicitly
balance performance degradation against
input fidelity at the perturbation-
configuration level.

\subsection{Perturbation-level robustness benchmarks}
\label{sec:plt}

For vision, ImageNet-C~\cite{imagenetc}
benchmarks robustness under 19 fixed
corruption types applied independently
at five severity levels.
It provides a rigorous single-perturbation
baseline but cannot capture interactions
between co-occurring corruptions.
For natural language,
TextAttack~\cite{morris2020textattack} provides a
unified framework for word-level adversarial
transformations on NLP models, covering
substitution, insertion, and deletion.
CheckList~\cite{ribeiro2020beyond} defines capability
tests across linguistic categories such
as negation, vocabulary, and robustness
to typos, with each capability tested
largely in isolation.
For code,
ReCode~\cite{wang2023recode} evaluates code
generation models 
under individual perturbation families
such as butterfingers, character case,
whitespace, and newline perturbations,
measuring pass@$k$ per family independently.
These benchmarks provide the perturbation
families used in our evaluation, but
they evaluate families independently
rather than searching over compositional
perturbation configurations under a budget.

\subsection{Configuration-level compositional testing}
\label{sec:clt}

Chandrasekaran et al.~\cite{chandrasekaran2021combinatorial}
showed that combinations of perturbations
can expose failures that are not observable
under isolated testing.
CIT4DNN~\cite{dola2024cit4dnn} addresses
the resulting combinatorial space through
combinatorial interaction testing in a
compressed latent space.
TestifAI~\cite{testifai2026} constructs
multi-perturbation spaces and estimates
higher-order robustness behaviour from
lower-order perturbation evaluations,
motivating the configuration space we
study.
With $n$ perturbation types and
$\ell$ discrete levels, however, the
space contains $\ell^n$ configurations.
For large models, evaluating each
configuration requires inference over
the full evaluation set, making
exhaustive evaluation costly.
What remains open is how to
explore this space under a limited
evaluation budget while prioritising
failures that are both severe and
high-fidelity.

\subsection{Multi-objective optimisation}
\label{sec:moo}

Multi-objective optimisation is common
in search-based software engineering,
where testing is formulated as a
trade-off between objectives such as
coverage, diversity, and fault
detection~\cite{miettinen1999,fraser2013whole,panichella2015reformulating}.
In AI testing,
DeepXplore~\cite{deepxplore} and
DLFuzz~\cite{guo2018dlfuzz}
combine multiple objectives, including
neuron coverage and behavioural divergence,
to generate failure-inducing inputs.
Most existing AI testing methods
apply multi-objective optimisation to
individual test inputs, rather than
perturbation configurations.

A common alternative is to scalarise
multiple objectives into a single
fitness score and apply single-objective
search.
However, scalarisation fixes the
degradation--fidelity trade-off, biasing
search toward one region of
the surface.
NSGA-II~\cite{deb2002fast} avoids this
by ranking configurations through
non-dominated sorting and preserving
diversity with crowding-distance selection.
\sys uses NSGA-II to search
the multi-perturbation configuration space,
optimising performance degradation and
input fidelity.

\SetKwComment{Comment}{$\triangleright$\ }{}
\SetCommentSty{textnormal}
\DontPrintSemicolon



\begin{algorithm}[t]
\small
\caption{\sys: NSGA-II search over \\
compositional perturbation configurations}
\label{alg:parrot}
$E \leftarrow \emptyset$\nllabel{alg:init-e}
\Comment*[r]{unique evaluated configurations}
$b \leftarrow 0$\;
$P \leftarrow \mu$ uniform samples
from $\Theta$\nllabel{alg:init-p} 

\ForEach{$\bm{\theta} \in P$}{ \nllabel{alg:init-loop}
  $\updelta(\bm{\theta}),\uprho(\bm{\theta})
  \leftarrow
  \textsc{Evaluate}(\bm{\theta})$\; \nllabel{alg:init-eval}
  $E \leftarrow E \cup \{\bm{\theta}\}$\; \nllabel{alg:init-add}
  $b \leftarrow b+1$\Comment*[r]{budget counter}
}

\While{$b < B$}{ \nllabel{alg:while}
  $m \leftarrow \min(\mu,\,B-b)$\nllabel{alg:remaining}\Comment*[r]{remaining budget}
  $Q \leftarrow
  \textsc{Variation}(P,\Theta,m)$\; \nllabel{alg:variation}

  \ForEach{$\bm{\theta} \in Q$}{ \nllabel{alg:q-loop}
  \If{$b = B$}{
    \textbf{break}\;
  }
  \If{$\bm{\theta} \notin E$}{ \nllabel{alg:if-new}
    $\updelta(\bm{\theta}),\uprho(\bm{\theta})
    \leftarrow
    \textsc{Evaluate}(\bm{\theta})$\; \nllabel{alg:q-eval}
    $E \leftarrow E \cup \{\bm{\theta}\}$\; \nllabel{alg:q-add}
  }
  $b \leftarrow b+1$\; \nllabel{alg:budget-inc}
}

  $\mathcal{F} \leftarrow$ rank
$P \cup Q$ by Pareto dominance\;
\nllabel{alg:rank}
$P \leftarrow$ select $\mu$
configs from $\mathcal{F}$ by rank and distance\;
\nllabel{alg:select}
}

$R \leftarrow$ sort $E$ by
rank, then distance\; \nllabel{alg:sort-output}
\Return{$R$}\; \nllabel{alg:return}
\end{algorithm}



\section{\sys Framework}
\label{sec:framework}

\sys identifies perturbation configurations
that expose severe model failures while
preserving input fidelity.
Given a model, a clean test
set, discrete perturbation space,
task-performance metric, fidelity metric,
and evaluation budget, \sys searches
for high-quality degradation--fidelity
trade-offs using NSGA-II, a
multi-objective evolutionary algorithm.
The result is a prioritised set
of configurations for practitioner
inspection.
We define the perturbation
configuration space
(\S\ref{sec:perturbation-space}), the
degradation and fidelity objectives
(\S\ref{sec:objectives}), and the
NSGA-II search procedure
(\S\ref{sec:nsga-search}).
 
\subsection{Perturbation Configuration Space}
\label{sec:perturbation-space}

We view perturbations as controlled
transformations intended to preserve task
semantics while shifting the input
distribution.
Let $\mathcal{D}$ be the clean
test set and let
$\{T_1,\ldots,T_n\}$ be $n$
perturbation types, such as blur,
noise, or whitespace corruption.
Each perturbation type $T_i$ is
applied at a discrete level
$\theta_i \in \{0,\ldots,\ell{-}1\}$,
where $\theta_i{=}0$ denotes no
perturbation.
A perturbation configuration
$\bm{\theta}=(\theta_1,\ldots,\theta_n)$
specifies one level per perturbation
type.
The resulting configuration space is
the discrete lattice
\[
  \Theta = \{0,\ldots,\ell{-}1\}^n .
\]
Applying configuration $\bm{\theta}$ to
an input $x$ composes the selected
perturbations:
\[
  \pi_{\bm{\theta}}(x)
  = T_n^{\theta_n} \circ \cdots
    \circ T_1^{\theta_1}(x),
\]
where $T_i^{\theta_i}$ denotes
perturbation type $T_i$ applied at
level $\theta_i$.
The \emph{perturbed test set $\mathcal{T}_{\bm{\theta}}$} for a configuration $\bm{\theta}$ is
\[
  \mathcal{T}_{\bm{\theta}}
  = \{\pi_{\bm{\theta}}(x) : x \in \mathcal{D}\}.
\]
In our experiments, $n{=}4$ and
$\ell{=}6$, yielding
$|\Theta| = 1{,}296$
perturbation configurations.

\subsection{Objectives}
\label{sec:objectives}

Each configuration $\bm{\theta}$ is
evaluated using two objectives.
The first is performance degradation:
how much task performance drops when
the test set is perturbed.
Let $\psi$ be a task-performance
metric, such as accuracy or
Robust Pass, e.g., RP$_5$@1~\cite{wang2023recode}.
We define the \emph{performance
degradation objective} $\updelta$ on
$\bm{\theta}$ as
\begin{equation}
  \label{eq:delta}
  \updelta(\bm{\theta})
  = \max\bigl(\psi(\mathcal{D}) {-}
    \psi(\mathcal{T}_{\bm{\theta}}),\,0\bigr).
\end{equation}
A higher $\updelta$ means greater
performance degradation.

The second objective is input fidelity:
how closely the perturbed inputs
resemble the originals.
Let $\phi$ be a modality-specific
fidelity score derived from a
similarity or distance metric.
We define \emph{input fidelity objective $\uprho$} on $\bm{\theta}$ as
\begin{equation}
  \label{eq:rho}
  \uprho(\bm{\theta}) =
  \phi(\mathcal{T}_{\bm{\theta}}, \mathcal{D}).
\end{equation}
A higher $\uprho$ means higher
input fidelity.

In our experiments, we instantiate
$\phi$ using SSIM
and KID for images,
and chrF
and BERT-F1
for text and code.
Because KID is a distance metric,
we invert it before normalisation
so that higher $\uprho$ consistently
indicates higher fidelity.
Both $\updelta$ and~$\uprho$ are
normalised to $[0,1]$ in
the experiments.

\subsection{NSGA-II Search}
\label{sec:nsga-search}

Algorithm~\ref{alg:parrot} summarises
the procedure.
\sys begins by sampling an initial
population $P$ of $\mu$ configurations
uniformly without replacement from
$\Theta$ and evaluating each on
$\updelta$ and $\uprho$
(lines~\ref{alg:init-p}--\ref{alg:init-add}).
\textsc{Evaluate}$(\bm{\theta})$
(line~\ref{alg:init-eval}) constructs
the perturbed test set
$\mathcal{T}_{\bm{\theta}}$, computes
the performance drop $\updelta(\bm{\theta})$
using Eq.~\ref{eq:delta}, and computes
the fidelity score $\uprho(\bm{\theta})$
using Eq.~\ref{eq:rho}.

\sys then repeats four steps 
until the proposal budget $B$
is exhausted.
It generates new candidates~$Q$
(line~\ref{alg:variation}), evaluates
any previously unseen candidates in
$Q$ (lines~\ref{alg:if-new}--\ref{alg:q-add}),
ranks configurations $P \cup Q$
by Pareto dominance
(line~\ref{alg:rank}), and selects
the next population $P$ by Pareto
rank and crowding distance
(line~\ref{alg:select}).

\textsc{Variation}$(P,\Theta,m)$
takes the current population $P$,
uses~$\Theta$ to enforce valid
severity levels, and produces up
to $m$ new candidates $Q$
(line~\ref{alg:variation}).
It applies simulated binary crossover
(SBX) and polynomial mutation to
parents sampled from $P$.
Because severity levels are integers,
fractional values produced by these
operators are rounded to the nearest
valid level in $\Theta$ using
a rounding repair step~\cite{blank2020pymoo}.

\sys ranks $P \cup Q$ using
Pareto dominance (line~\ref{alg:rank}).
A configuration $\bm{\theta}$ dominates
$\bm{\theta}'$ if it is at
least as good on both $\updelta$
and $\uprho$, and strictly better
on at least one.
Over the full configuration space,
the ground-truth Pareto front
$\mathcal{P}^*$ contains all configurations
in $\Theta$ that are dominated
by no other configuration in
$\Theta$.
\sys does not observe $\mathcal{P}^*$
during search; it approximates it
by applying non-dominated sorting to
the current candidate set $P \cup Q$,
partitioning candidates into successive
fronts.
Crowding distance estimates how isolated
a configuration is in the
$(\updelta,\uprho)$ objective space;
larger values indicate less crowded
regions of the trade-off surface.
The next population is filled from
the best-ranked fronts first.
When a front does not fit,
crowding-distance selection retains candidates
with larger crowding distance, preserving
diversity across the degradation--fidelity
trade-off (line~\ref{alg:select}).

The loop continues until the proposal
budget $B$ is exhausted.
At the end, \sys returns
$R$, the evaluated perturbation
configurations ranked by Pareto rank
and crowding distance.
This gives practitioners a prioritised
list of severe, high-fidelity
configurations for robustness inspection.
 
\paragraph{Hyperparameter selection.}

NSGA-II requires three search parameters,
which we tune by grid search:
\begin{inparaenum}[\it (i)]
\item
the crossover index $\eta_c$ controls
how far offspring spread from their
parents under SBX crossover: lower
values promote broader exploration, while
higher values favour local refinement.
We search over
$\eta_c\, {\in}\, \{1,5,10,15\}$;
\item
the mutation index $\eta_m$ controls
the step size of polynomial mutation,
with lower values producing larger
steps.
We search over
$\eta_m\,{\in}\,\{1,5,10,15\}$;
\item
the population size $\mu$ determines
how many configurations are maintained
per generation; larger populations preserve
more Pareto-front diversity but consume
more budget per generation.
We search over
$\mu \,{\in}\,\{20,50,80,110\}$.
\end{inparaenum}

This grid contains 64
settings,
each
evaluated on all four benchmarks
under two fidelity metrics per
benchmark: SSIM and KID for
images, and chrF and BERT-F1
for text and code.
With 10 random seeds, this
yields $5{,}120$ runs.
We select the setting that maximises
the number of benchmark--metric settings
reaching Recall@$\mathcal{P}^*{\geq}0.999$,
breaking ties by worst-case recall.
This selects $\eta_c{=}1$,
$\eta_m{=}5$, and $\mu{=}110$.

\begin{table*}[!t]
\centering\small
\resizebox{\textwidth}{!}{
\begin{tabular}{@{}llrllllr@{}}
\toprule
& Dataset $\mathcal{D}$ & Size & Model & Perturbations & Fidelity $\phi$ & Task $\psi$ & Clean $\psi(\mathcal{D})$ \\
\midrule
\imagenetSymb & Tiny-ImageNet & 10,000 & CaiT-S36 &
  speckle noise, glass blur, brightness, pixelate &
  KID,\ SSIM & Accuracy & 86.7\% \\

\qqpSymb & QQP & 1,000 & RoBERTa-base &
  synonym, typo, contraction, punctuation &
  BERT-F1,\ chrF & Accuracy & 91.2\% \\
\recodeSymb & HumanEval & 164 & CodeGen-2B-mono &
  butterfingers, char case, whitespace, newline &
  BERT-F1,\ chrF & RP$_5$@1 & 23.2\% \\
\mbppSymb & MBPP& 974 & CodeGen-2B-mono&
  butterfingers, char case, whitespace, newline &
  BERT-F1,\ chrF & RP$_5$@1 & 31.9\% \\
\bottomrule
\end{tabular}
}
\caption{Benchmarks, datasets and models,
perturbations, fidelity metrics
$\phi$, task-performance metrics $\psi$,
and clean-set performance $\psi(\mathcal{D})$.}
  \label{tab:benchmarks}
\end{table*}

\begin{figure*}[t]
\centering
\resizebox{0.8\textwidth}{!}{%
\begin{tabular}{@{}cccc@{}}
  \begin{subfigure}[t]{0.24\textwidth}
    \includegraphics[width=\linewidth]{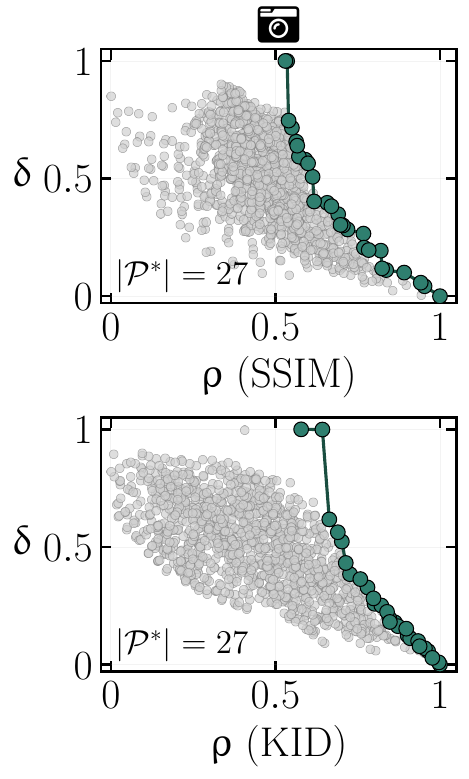}
  \end{subfigure} &
  \begin{subfigure}[t]{0.24\textwidth}
    \includegraphics[width=\linewidth]{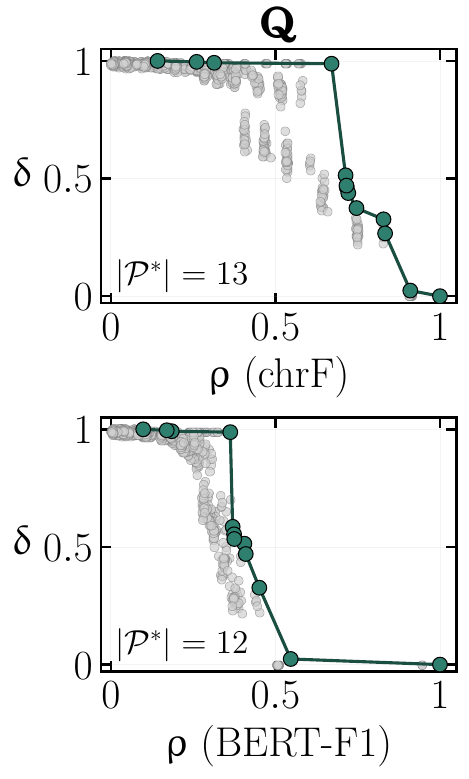}
  \end{subfigure} &
  \begin{subfigure}[t]{0.24\textwidth}
    \includegraphics[width=\linewidth]{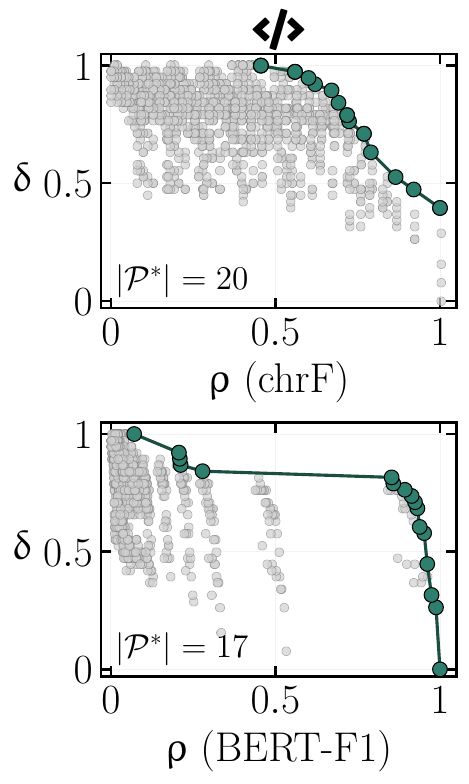}
  \end{subfigure} &
  \begin{subfigure}[t]{0.24\textwidth}
    \includegraphics[width=\linewidth]{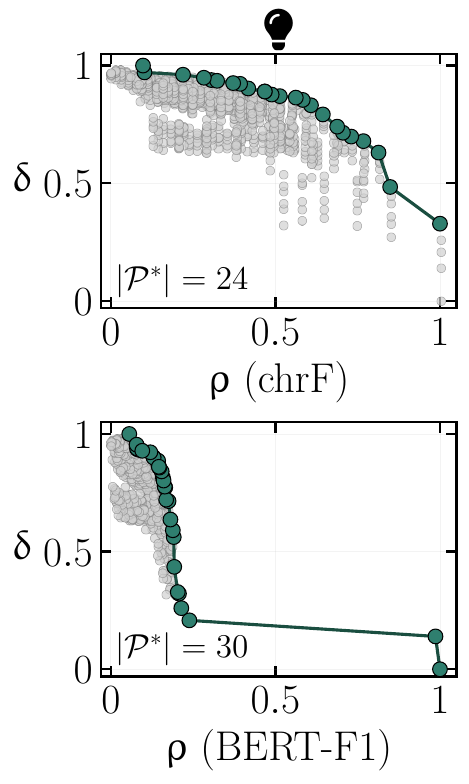}
  \end{subfigure}
\end{tabular}}
\caption{$\mathcal{P}^*$ across benchmarks and fidelity metrics.
Each panel shows the full configuration space (grey) and Pareto front (teal);
$|\mathcal{P}^*|$ gives the front size per benchmark--metric pair.
Pareto fronts contain configurations with one to four active perturbation
dimensions. Counts by order 1--4 are:
\imagenetSymb~SSIM $27$ (9,12,4,2); KID $27$ (4,7,14,2);
\qqpSymb~chrF $13$ (2,7,2,2); BERT-F1 $12$ (2,7,1,2);
\recodeSymb~chrF $20$ (2,3,12,3); BERT-F1 $17$ (3,9,5,0);
\mbppSymb~chrF $24$ (1,5,13,5); BERT-F1 $30$ (7,12,5,6). See Appendix for details.}
\label{fig:pareto_all_datasets}
\end{figure*}
 

\begin{figure*}[!t]
\centering
\resizebox{0.85\textwidth}{!}{%
\begin{tabular}{c@{\hspace{8pt}}c}
  \includegraphics[width=0.49\textwidth]{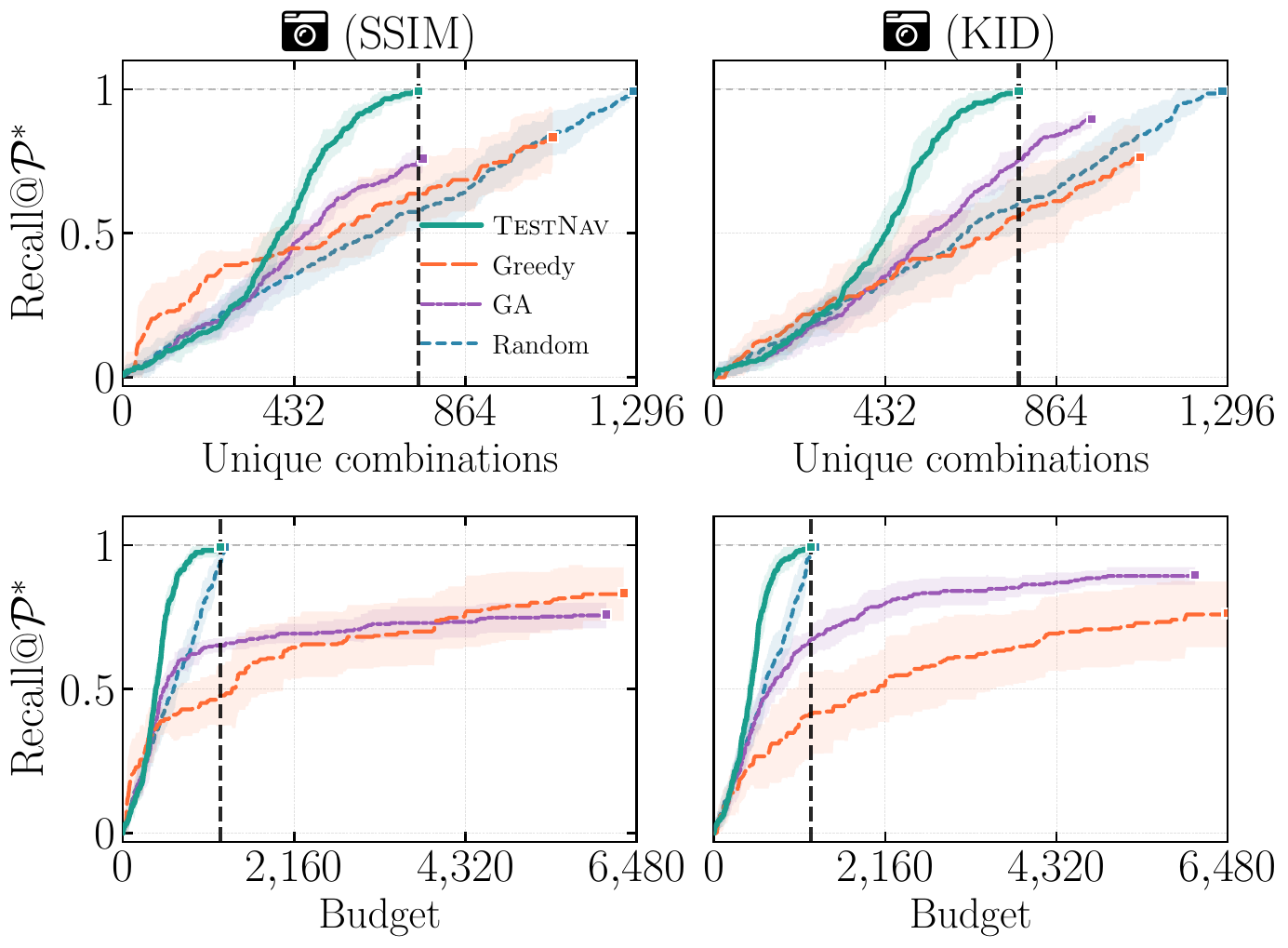} &
  \includegraphics[width=0.49\textwidth]{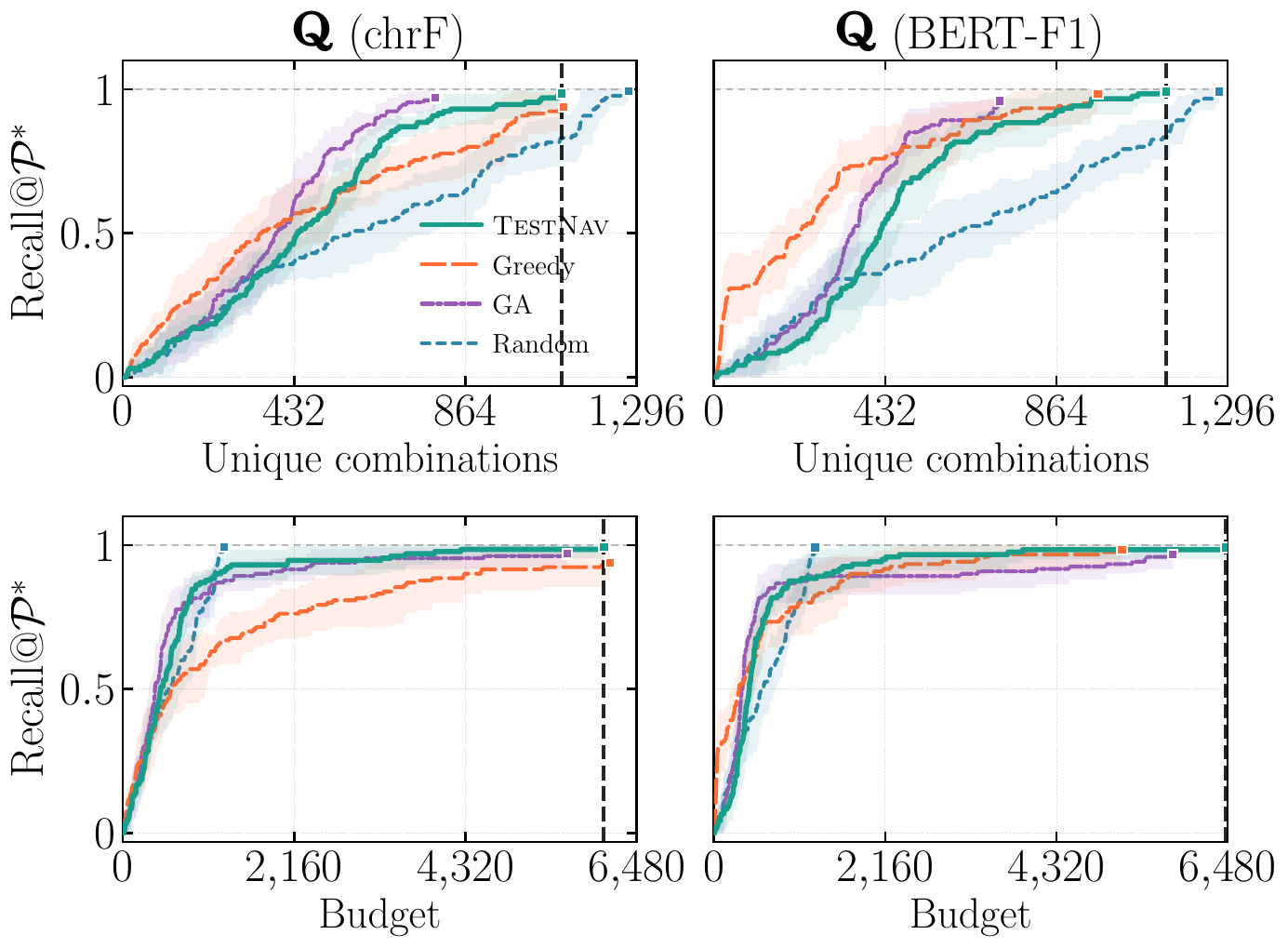} \\[8pt]
  \includegraphics[width=0.49\textwidth]{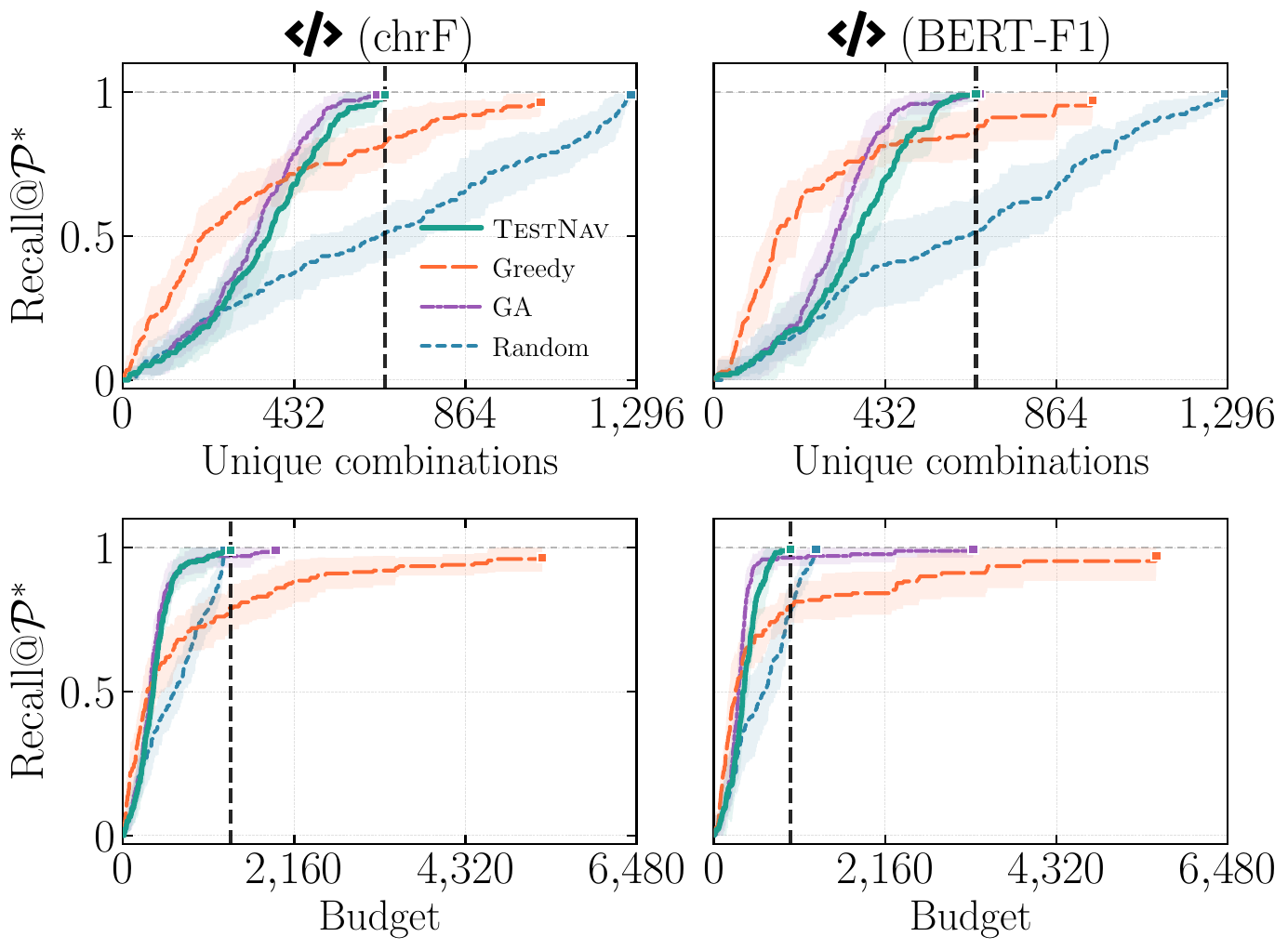} &
  \includegraphics[width=0.49\textwidth]{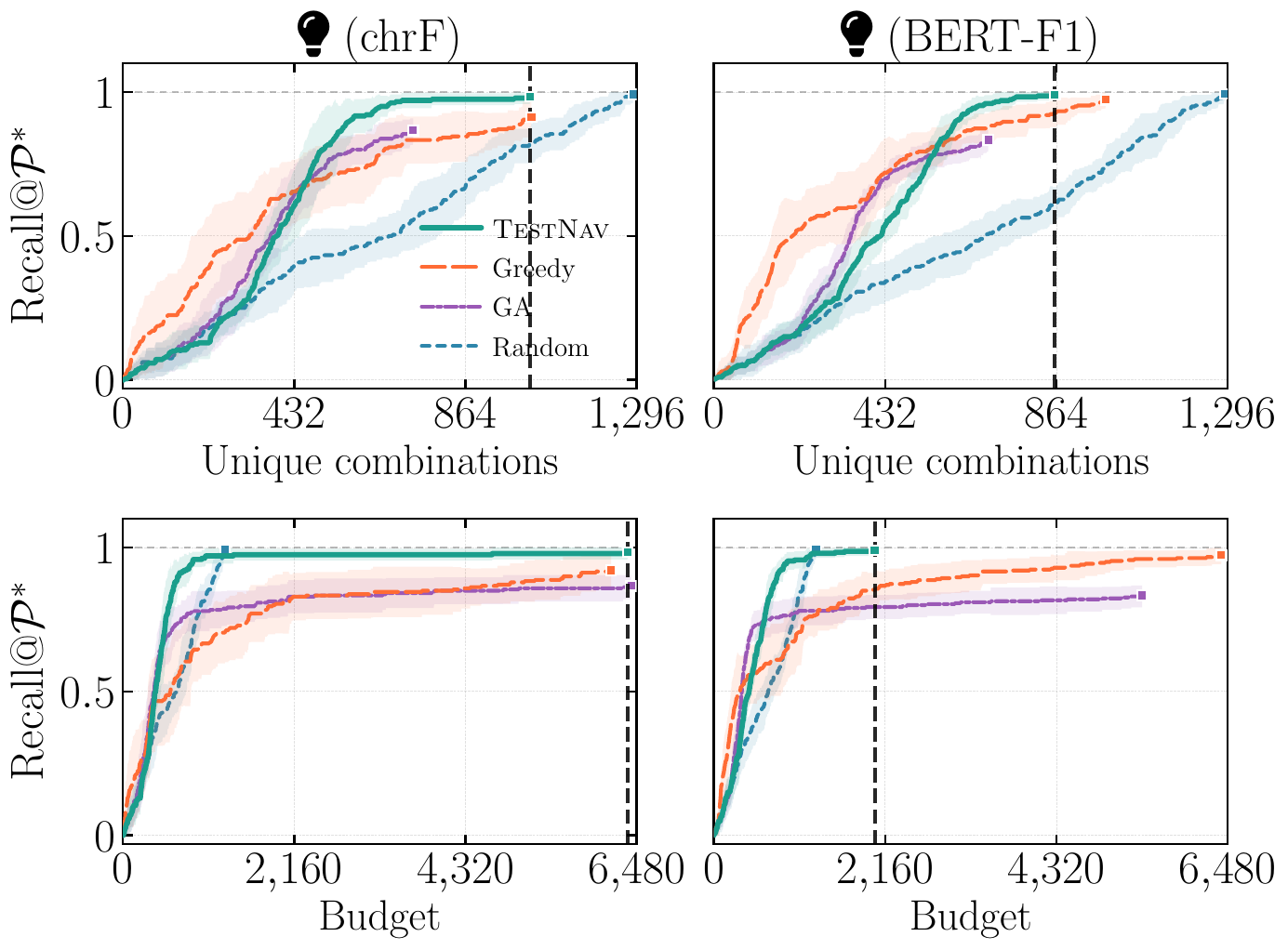}
\end{tabular}}
\caption{Recall@$\mathcal{P}^*$ across
benchmarks and fidelity metrics.
For each benchmark, the top row uses
unique configurations as the $x$-axis,
and the bottom row uses evaluation
budget.
Shaded bands show $\pm1$ standard
deviation over 10 seeds. The vertical line indicates the number of unique perturbation combinations (top) and total budget (bottom) required by \textsc{TestNav} to discover all Pareto-optimal configurations~$\mathcal{P}^*$; endpoint markers ($\blacksquare$) show the highest Recall@$\mathcal{P}^*$ achieved by each method..
}
\label{fig:main_results}
\end{figure*}


\begin{table}[t]
\centering
\setlength{\tabcolsep}{4.5pt}
\huge
\resizebox{\columnwidth}{!}{%
\begin{tabular}{lcccccccc}
\toprule
& \multicolumn{2}{c}{\imagenetSymb}
& \multicolumn{2}{c}{\qqpSymb}
& \multicolumn{2}{c}{\recodeSymb}
& \multicolumn{2}{c}{\mbppSymb} \\
\cmidrule(lr){2-3}
\cmidrule(lr){4-5}
\cmidrule(lr){6-7}
\cmidrule(lr){8-9}
Method
  & SSIM & KID
  & chrF & BERT-F1
  & chrF & BERT-F1
  & chrF & BERT-F1 \\
\midrule
\textbf{\sys}
  & \textbf{0.704}& \textbf{0.686}
  & \underline{0.646} & 0.651
  & 0.729 & 0.730
  & \textbf{0.697} & \underline{0.690} \\
Greedy Search
  & \underline{0.557} & 0.464
  & 0.638 & \textbf{0.782}
  & \underline{0.742} & \textbf{0.783}
  & \underline{0.681} & \textbf{0.751} \\
Genetic Algorithm
  & 0.540 & \underline{0.552}
  & \textbf{0.693} & \underline{0.707}
  & \textbf{0.753} & \underline{0.767}
  &  0.640 & 0.631 \\
Random Search
  & 0.510 & 0.503
  & 0.526 & 0.514
  & 0.502 & 0.522
  & 0.514 & 0.490 \\
\noalign{\vskip 4pt}
\bottomrule
\end{tabular}
}
\caption{AUC-Recall@$\mathcal{P}^*$ for
\sys and search baselines across all
benchmarks and fidelity metrics.
\textbf{Bold} indicates the best score
and \underline{underline} indicates 
the second-best
score per column.}
\label{tab:auc_combined}
\end{table}
\begin{figure*}[!t]
\centering
\resizebox{0.85\textwidth}{!}{%
\begin{tabular}{c@{\hspace{8pt}}c}
  \includegraphics[width=0.49\textwidth]{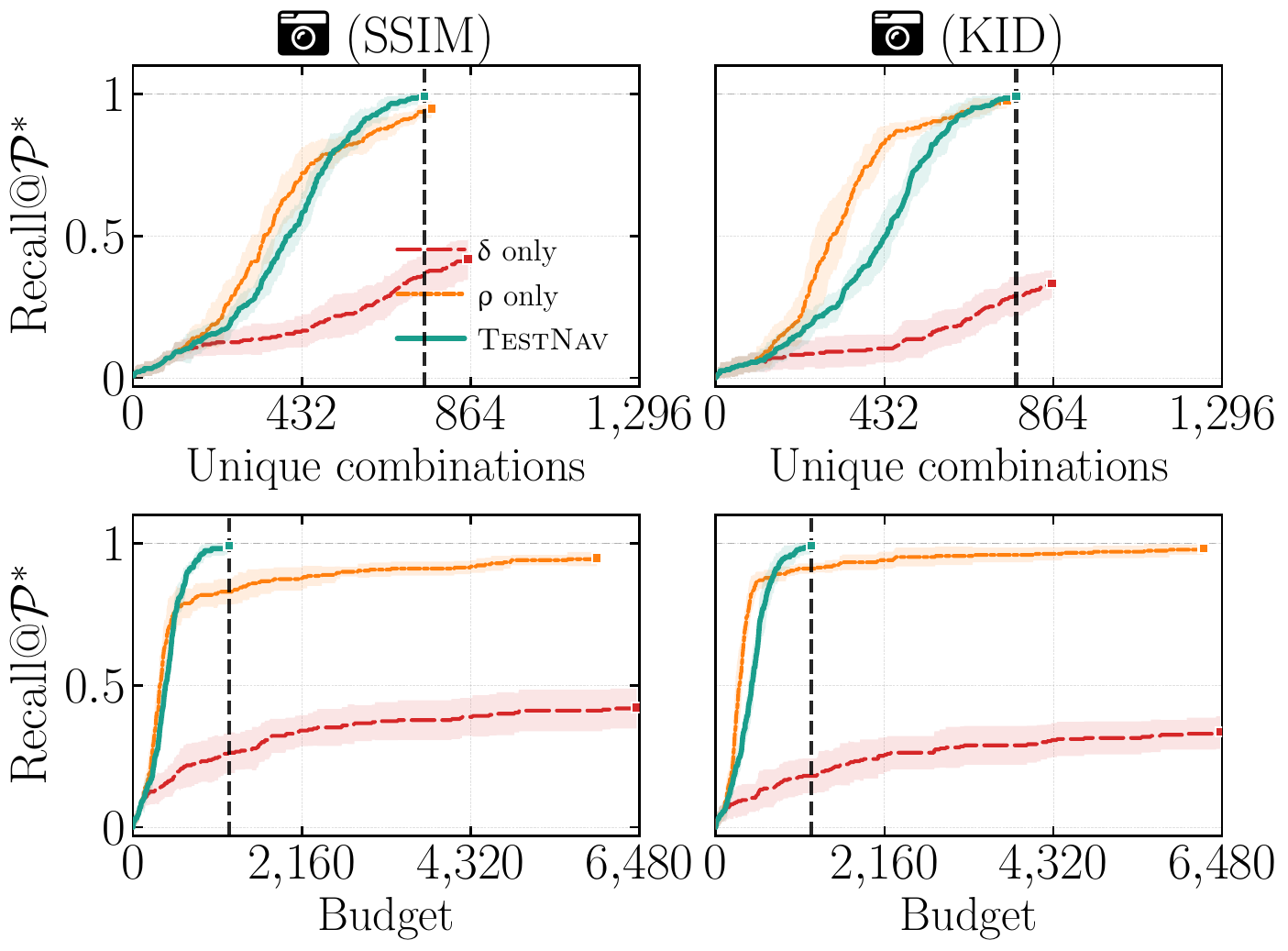} &
  \includegraphics[width=0.49\textwidth]{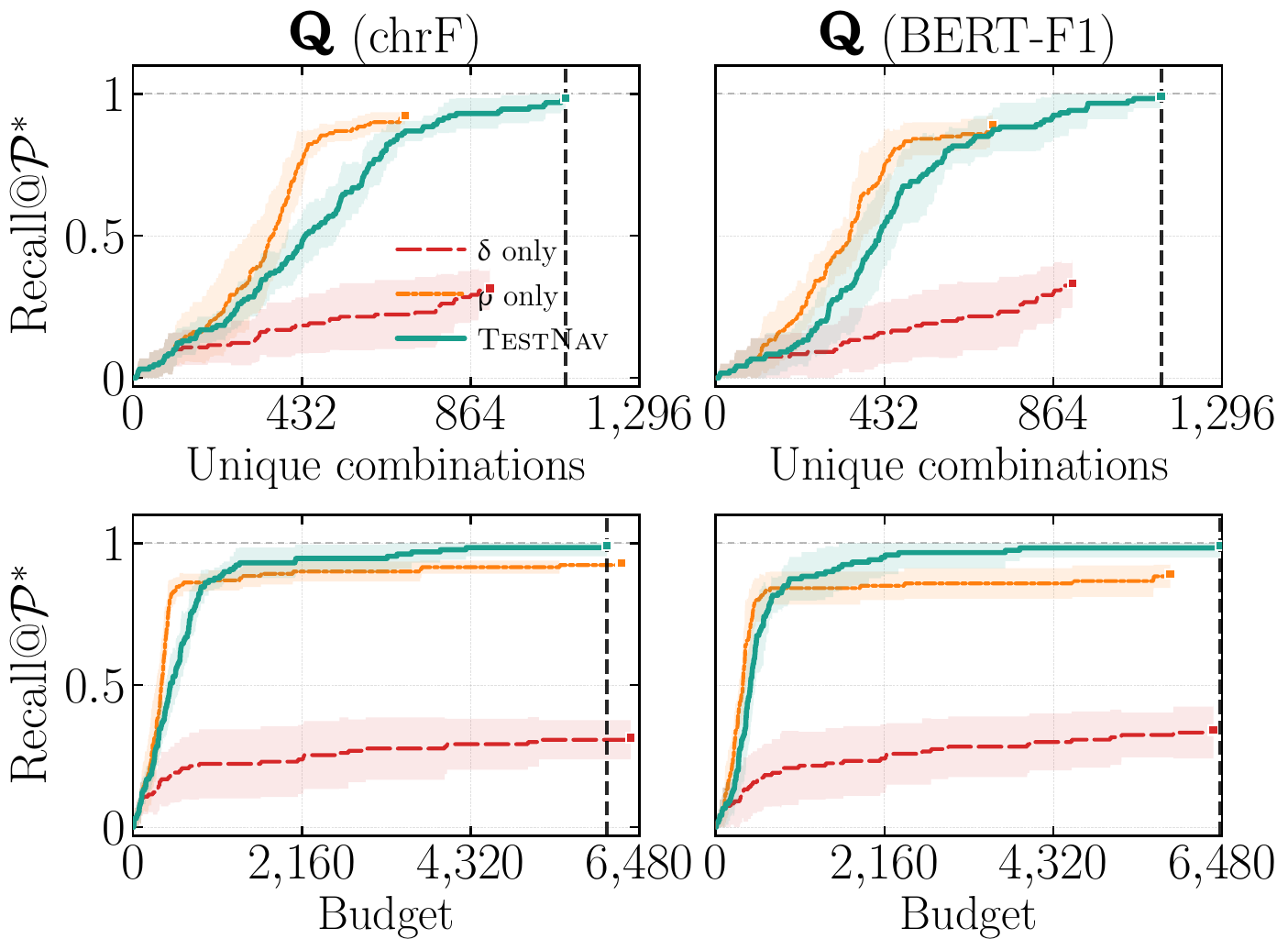} \\[8pt]
  \includegraphics[width=0.49\textwidth]{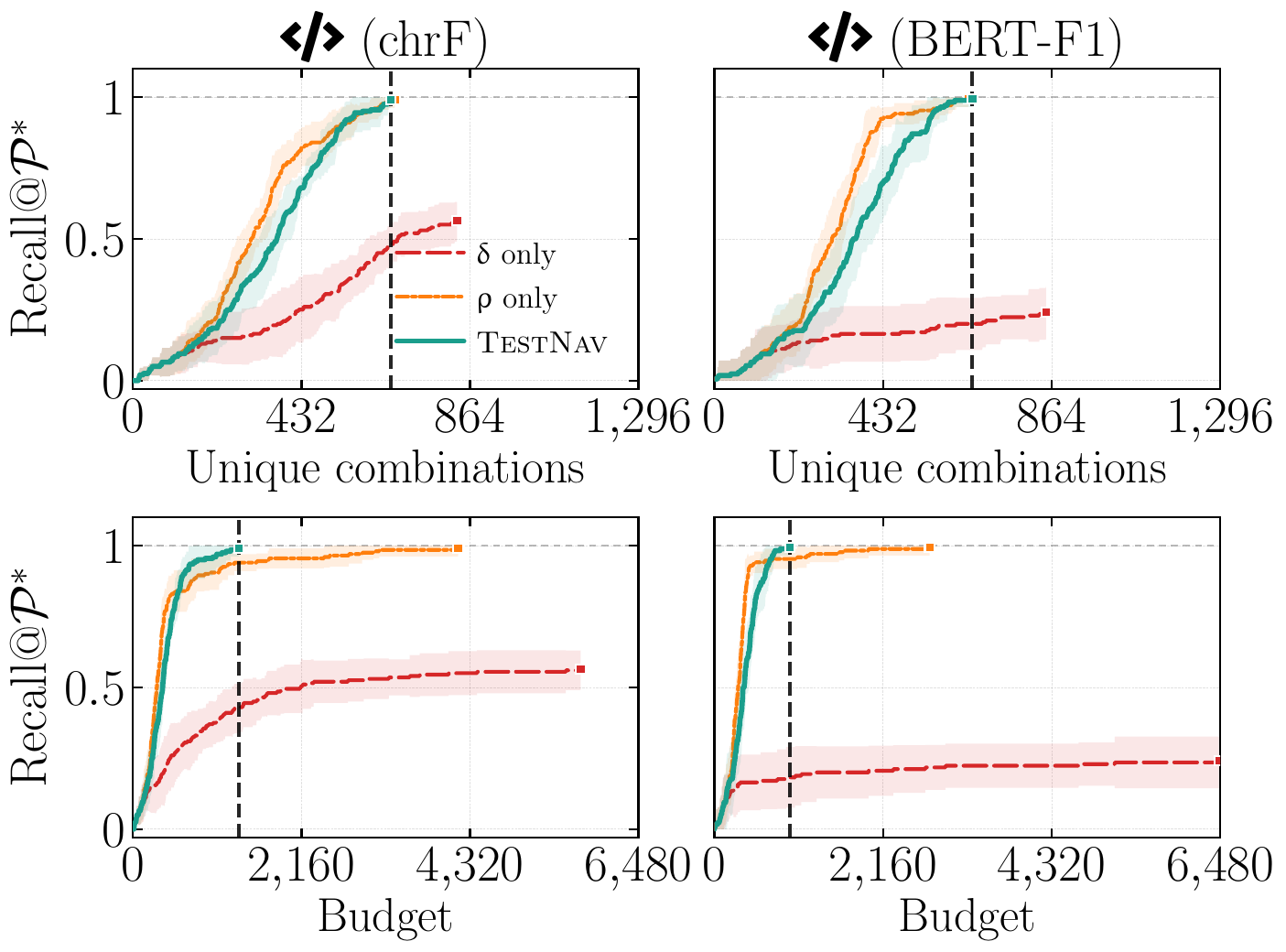} &
  \includegraphics[width=0.49\textwidth]{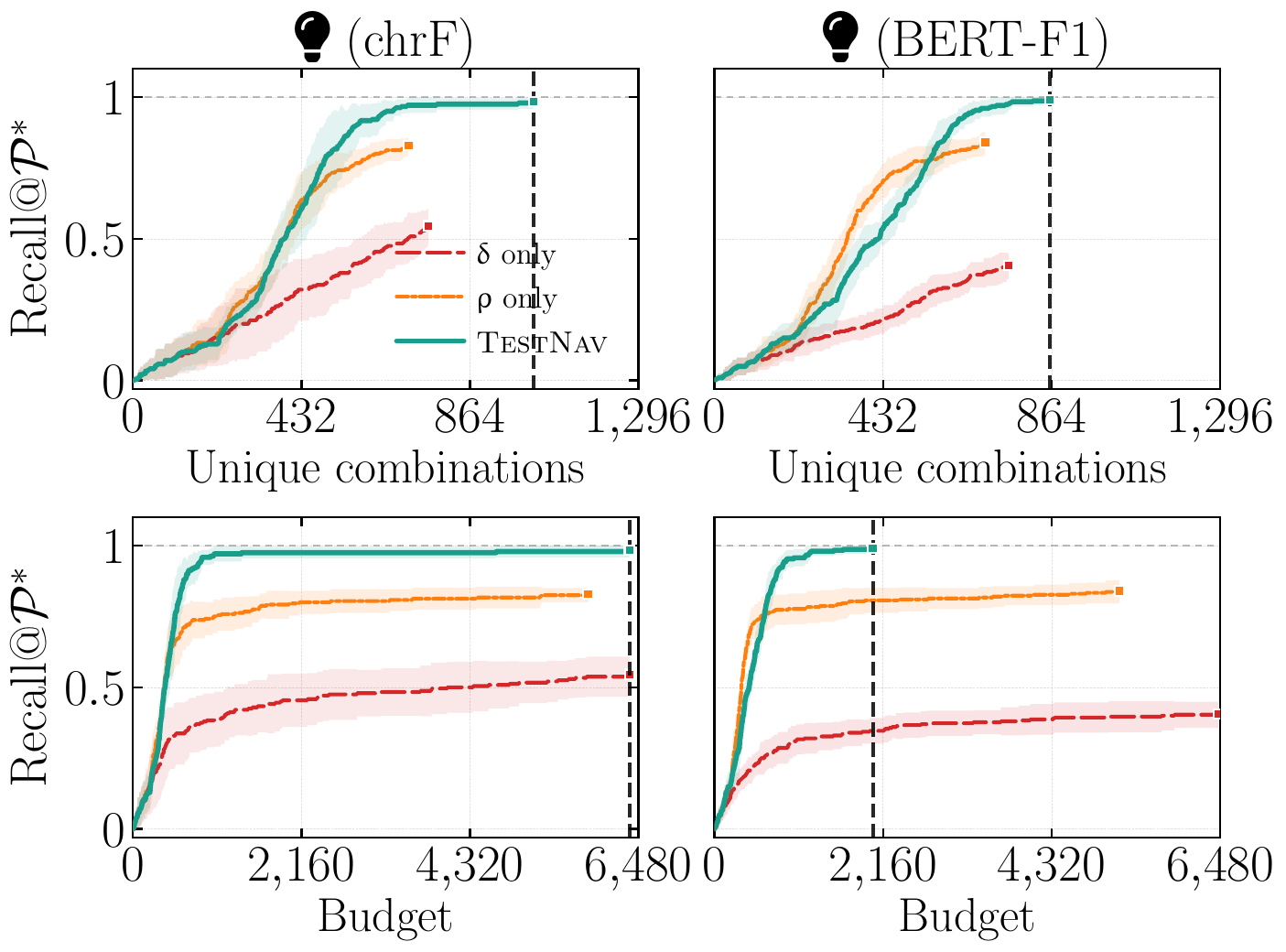}
\end{tabular}}
\caption{Multi-objective versus single-objective
search across benchmarks and fidelity metrics.
Curves show Recall@$\mathcal{P}^*$
over unique configurations in the top row
and evaluation budget in the bottom row;
bands show $\pm1$ standard deviation
over 10 seeds.}
\label{fig:ablation}
\end{figure*}
\begin{figure}[!t]
\centering
\resizebox{0.425\textwidth}{!}{%
  \includegraphics[width=\linewidth]{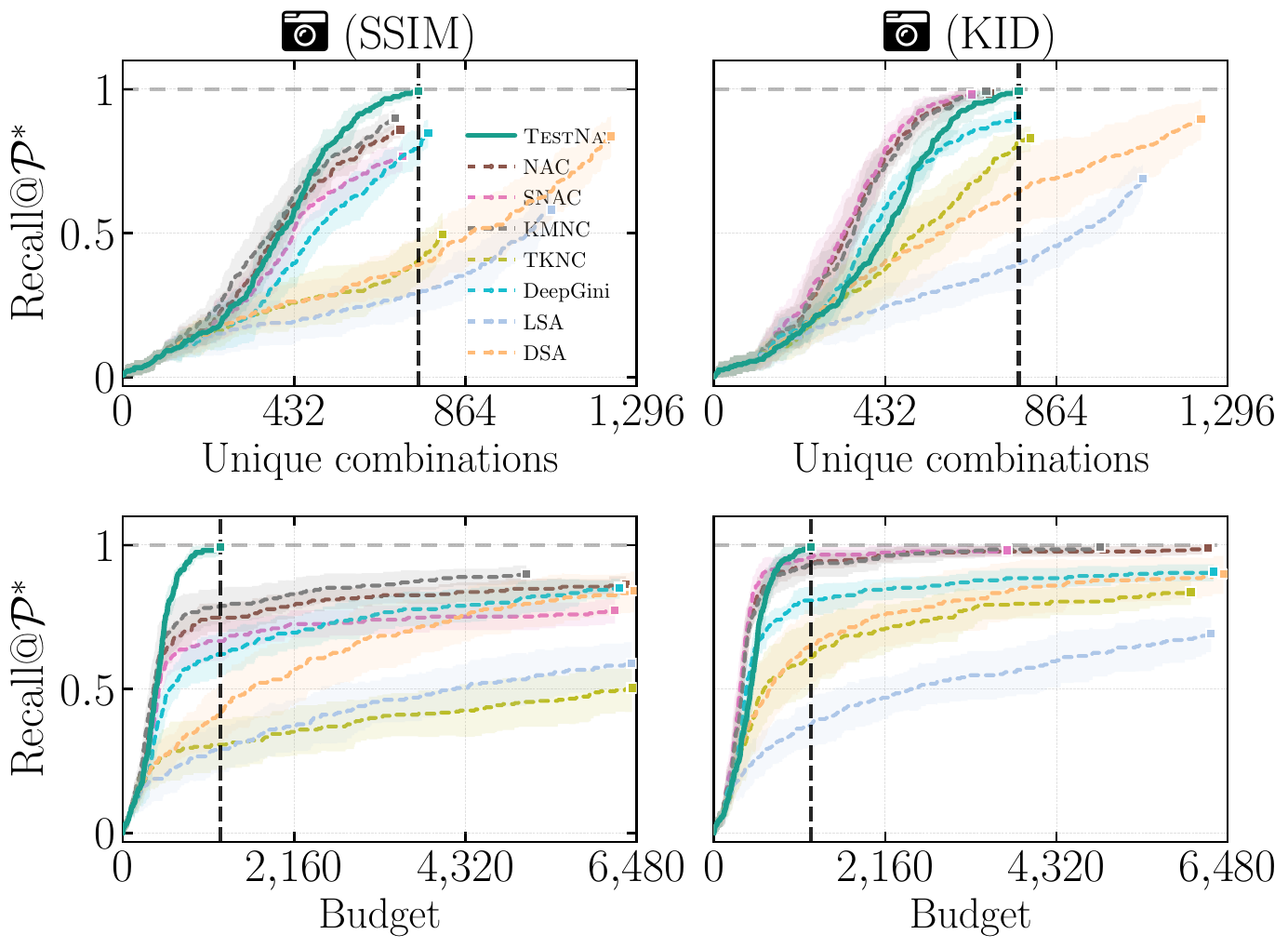}}
\caption{Input-level metrics versus \sys on
{\footnotesize \imagenetSymb}.
Curves show Recall@$\mathcal{P}^*$
over unique configurations (top) and
evaluation budget (bottom), using SSIM
(left) and KID-derived fidelity (right).}
\label{fig:coverage}
\end{figure}

\section{Evaluation}
\label{sec:evaluation}

We address three research questions:
\begin{inparaenum}[\it (i)]
\item
whether \sys recovers $\mathcal{P}^*$
more efficiently than non-Pareto
baselines (\S\ref{sec:rq1});
\item
whether single-objective search can
recover $\mathcal{P}^*$ (\S\ref{sec:rq2}); and
\item
whether input-level test metrics can
proxy the degradation--fidelity trade-off
(\S\ref{sec:rq3}).
\end{inparaenum}

\subsection{Experimental Setup}
\label{sec:setup}

\paragraph{Benchmarks.}
Table~\ref{tab:benchmarks} summarises
the four benchmarks, each defining
four perturbation dimensions with six
severity levels ($0$--$5$), where
level~0 denotes no perturbation:

\begin{inparaenum}[\it (i)]
\item
\emph{Image classification}
({\footnotesize \imagenetSymb}).
We evaluate CaiT-S36~\cite{touvron2021going}
on 10{,}000 Tiny-ImageNet validation
images~\cite{deng2024tinyimagenet} using
four ImageNet-C corruptions: speckle
noise, glass blur, brightness, and
pixelate~\cite{imagenetc,imagecorruptions}.
Levels~1--5 follow the standard
ImageNet-C severity scale.

\item
\emph{Paraphrase detection} 
({\footnotesize \qqpSymb}).
We evaluate {\small $\textsf{RoBERTa}_{\textsf{base}}$}
on 1{,}000 Quora Question Pairs
(QQP), where each pair is labelled
for semantic equivalence~\cite{wang2018glue}.
We use TextAttack~\cite{morris2020textattack}
to apply synonym replacement, typos,
contractions, and punctuation perturbations.
Levels~1--5 correspond to applying
one to five edits.

\item
\emph{Code generation}.
We evaluate 
{\small \textsf{CodeGen-2B-mono}}
on HumanEval ({\footnotesize \recodeSymb}) and MBPP ({\footnotesize \mbppSymb})~\cite{humaneval,austin2021program,nijkamp2022codegen}.
Following ReCode~\cite{wang2023recode}, we
use butterfingers, character-case, whitespace,
and newline perturbations,
but apply them compositionally.
The same six severity levels are
used for both datasets.\footnote{For levels~1--5, butterfingers
is applied with probability
$p\in\{0.05,
\allowbreak
0.1,
\allowbreak
0.15,
\allowbreak
0.2,
\allowbreak
0.25\}$,
and character-case with
$p\in\{0.1,
\allowbreak
0.2,
\allowbreak
0.35,
\allowbreak
0.5,
\allowbreak
0.7\}$;
whitespace characters are added and
deleted with
$(p_{\mathrm{add}},p_{\mathrm{del}})
\allowbreak
\in
\allowbreak
\{(0.1,0.05),
\allowbreak
(0.15,0.07),
\allowbreak
(0.2,0.1),
\allowbreak
(0.25,0.12),
\allowbreak
(0.3,0.15)\}$;
and newline insertion adds
$n\in\{1,2,3,4,5\}$ lines.}
\end{inparaenum}

\paragraph{Task performance and fidelity metrics.}
For each configuration $\bm{\theta}$,
we compute task performance
$\psi(\mathcal{T}_{\bm{\theta}})$ and
input fidelity
$\phi(\mathcal{T}_{\bm{\theta}},\mathcal{D})$
at the configuration level.
Task performance is accuracy for
{\footnotesize \imagenetSymb} 
and 
{\footnotesize \qqpSymb}, 
and Robust Pass RP$_5$@1 for
{\footnotesize \recodeSymb}
and
{\footnotesize \mbppSymb}~\cite{wang2023recode}.

For input fidelity, we use SSIM
and KID for 
{\footnotesize \imagenetSymb}, 
and chrF and BERT-F1 for 
{\footnotesize \qqpSymb},
{\footnotesize \recodeSymb}, and
{\footnotesize \mbppSymb}.
SSIM is computed per perturbed
image against its clean original,
then averaged over the test set.
KID is computed once per configuration
between the clean and perturbed image
distributions with subset size 1{,}000;
we negate it before normalisation so
that higher values indicate greater
fidelity.
chrF is computed over the full
set of perturbed and clean inputs.
BERT-F1 is computed using
{\small $\textsf{RoBERTa}_{\textsf{base}}$}
for {\footnotesize \qqpSymb} and
{\small $\textsf{CodeBERT}_{\textsf{base}}$}~\cite{feng2020codebert}
for {\footnotesize \recodeSymb} and
{\footnotesize \mbppSymb}.
All fidelity scores are normalised
to $[0,1]$, with higher values
indicating greater input fidelity.\footnote{We
use the TorchMetrics library for SSIM,
KID, and BERT-F1~\cite{detlefsen2022torchmetrics},
and \textsc{SacreBLEU} for
chrF~\cite{post2018sacrebleu}.}

\paragraph{Ground-truth Pareto front.}
$\mathcal{P}^*$ was computed exhaustively
by evaluating all $|\Theta|{=}1{,}296$
configurations and retaining those
not dominated under
$(\updelta,\uprho)$~\cite{deb2002fast}.
It serves as the fixed ground truth
against which all methods are evaluated.
Figure~\ref{fig:pareto_all_datasets} shows
the resulting fronts for each benchmark
and fidelity metric.
The fronts contain configurations with
one to four active perturbations, showing
that Pareto-optimal failures are not
limited to single perturbations.

\paragraph{Search metrics.}
We set the evaluation budget to
$B\,{=}\,5{\times}|\Theta|{=}6{,}480$
configuration proposals.
A proposal evaluates a configuration
on the full perturbed test set unless
the configuration has already been seen,
in which case cached $\updelta$ and
$\uprho$ values are reused.
We set $B$ larger than $|\Theta|$
only to observe full search traces,
including convergence behaviour and repeated
proposals; efficiency is
measured by the number of unique
configurations required to recover
$\mathcal{P}^*$ and by the recall
trajectory defined below.
We also track $u$, the number
of unique configurations evaluated so far;
$u \leqslant B$ because search may
revisit configurations.

We measure recovery of the ground-truth
Pareto front using Recall@$\mathcal{P}^*$.
Let $E_u \subseteq \Theta$ be
the set of the first $u$ unique
configurations evaluated by a method.
Then
\[
  \mathrm{Recall@}\mathcal{P}^*(u)
  =
  \frac{|E_u \cap \mathcal{P}^*|}
       {|\mathcal{P}^*|}.
\]
A value of 1 means that
all Pareto-optimal configurations have
been found.
Steeper recall curves indicate
earlier discovery of $\mathcal{P}^*$.

AUC-Recall summarises the full recall
trajectory over $u=1,\ldots,|\Theta|$
unique configurations.
Let $r_i=\mathrm{Recall@}\mathcal{P}^*(i)$.
Then
\[
  \mathrm{AUC\mbox{-}Recall@}\mathcal{P}^*
  =
  \frac{1}{|\Theta|}
  \sum_{i=1}^{|\Theta|-1}
  \frac{r_i+r_{i+1}}{2}.
\]
Higher AUC-Recall indicates earlier
discovery of $\mathcal{P}^*$ under
the same evaluation budget.

\paragraph{Execution protocol.}
All methods are evaluated under the
same proposal budget $B$ and run
with 10 independent seeds on a
GPU cluster at the Massachusetts Green
High Performance Computing Center (MGHPCC).

\subsection{Does \sys recover
  $\mathcal{P}^*$ more efficiently than non-Pareto baselines?}
\label{sec:rq1}





\sys uses Pareto-guided bi-objective
selection over performance degradation
and input fidelity.
We compare it against three baselines:
objective-free random search, local
scalar search, and scalar genetic
search.
For the scalar baselines, we use
$s(\bm{\theta}){=}\min(\updelta(\bm{\theta}),
\uprho(\bm{\theta}))$, so a configuration
receives a high score only when
both degradation and fidelity are high.
The baselines are:
\begin{inparaenum}[\it (i)]
\item
\emph{Random Search} evaluates
configurations in $\Theta$ in a
uniformly random order, visiting each
configuration exactly once;
\item
\emph{Greedy Search} starts from a
random configuration and repeatedly moves
to the highest-scoring neighbour under
$s$, where neighbours differ by
$\pm1$ in one severity dimension,
restarting when no neighbour improves;
\item
\emph{Genetic Algorithm} uses the same
population size, crossover, mutation, and
duplicate-elimination settings as \sys,
but optimises the scalar score $s$
rather than selecting by Pareto rank
and crowding distance.
\end{inparaenum}

Figure~\ref{fig:main_results} reports
Recall@$\mathcal{P}^*$ over 10 seeds.
The top-row $x$-axis shows unique
configurations visited and the bottom-row
$x$-axis shows the corresponding consumed
evaluation budget; shaded regions show
$\pm1$ standard deviation.
Table~\ref{tab:auc_combined} summarises
the corresponding AUC-Recall values.

\paragraph{Pareto search helps on broad fronts.}
Using SSIM for vision and chrF
for language/code, 
\sys achieves the
highest AUC-Recall on \imagenetSymb and \mbppSymb, while Genetic Algorithm is highest on \qqpSymb and \recodeSymb
(Table~\ref{tab:auc_combined}).
\sys fully recovers $\mathcal{P}^*$ 
up to $2.15\times$
earlier than
baselines that also achieve
$\text{Recall@}\mathcal{P}^*{=}1$,
evaluating 35.8\%--89.3\% of $\Theta$.
\sys achieves AUC
0.704
on {\footnotesize \imagenetSymb}, 
which is 38\% above
Random Search (0.510) and 26\%
above Greedy Search (0.557).
These gains suggest that, when the
Pareto front is broad, diversity-preserving
selection helps recover degradation--fidelity
trade-offs that scalar search misses.

\paragraph{Greedy search is competitive on compact fronts.}
With KID-derived fidelity for vision
and BERT-F1 for language/code, Greedy
Search achieves higher AUC than \sys on
{\footnotesize \qqpSymb}, {\footnotesize \recodeSymb},
and {\footnotesize \mbppSymb}
(Table~\ref{tab:auc_combined}).
Figure~\ref{fig:pareto_all_datasets} suggests
why: under these metrics,
$\mathcal{P}^*$ is small and concentrated
in the $(\updelta,\uprho)$ space,
so the scalar score
$s(\bm{\theta})$ can recover much of
$\mathcal{P}^*$ without exploring broadly.
When $\mathcal{P}^*$ is larger or
more spread out, as under SSIM and
chrF, scalar search is less effective,
as it tends to concentrate on
one region of the trade-off surface.
\sys mitigates this collapse through
crowding-distance selection, which encourages
the population to spread across the
degradation--fidelity surface.

\subsection{Can single-objective search recover
  $\mathcal{P}^*$?}
\label{sec:rq2}

Figure~\ref{fig:ablation} compares
\sys against two single-objective
ablations using identical evolutionary
operators:
\emph{$\updelta$ only} maximises
degradation without an input
fidelity objective; \emph{$\uprho$ only}
maximises input fidelity without a
degradation objective.

\paragraph{Single-objective search recovers less of the front.}
Both single-objective ablations recover
substantially less of $\mathcal{P}^*$.
The $\updelta$-only condition favours
high-degradation
configurations,
often at the cost
of input fidelity; its AUC ranges
from 0.185 to 0.384, up to 0.545
below \sys.
The $\uprho$-only condition favours
high-fidelity configurations, but lacks
a signal for model failure; its AUC
ranges from 0.612 to 0.771, within 0.085 of \sys.
Both show wider variance
bands than \sys, suggesting 
lower stability
across seeds.
\sys maintains spread across the
degradation--fidelity surface through
crowding-distance selection.
Together, these results show that
both objectives are necessary for
Pareto-front recovery and stability.

\subsection{Can input-level metrics proxy the trade-off?}
\label{sec:rq3}

Figure~\ref{fig:coverage} evaluates seven
input-level baselines introduced in
\S\ref{sec:ilt} on 
{\footnotesize \imagenetSymb}: 
NAC, SNAC, KMNC,
TKNC, DeepGini, LSA, and DSA.
Each metric is used as the sole
search objective; the ground truth is
the corresponding bi-objective Pareto front.



\paragraph{Input-level signals are insufficient.}
No input-level metric matches \sys
in final
Recall@$\mathcal{P}^*$.
Under SSIM, \sys reaches 0.993,
with KMNC next best,
at 0.900 (9\% lower); 
TKNC (0.496) and LSA (0.581)
are weakest.
Under KID-derived fidelity, \sys again
reaches 0.993, matched by KMNC (0.993).
Input-level metrics
measure activation coverage,
prediction uncertainty, or distributional
novelty on individual inputs, but
not where a perturbation
configuration lies on the
$(\updelta,\uprho)$ surface.
Thus, maximising them can
prioritise novel or
activation-diverse
configurations 
without 
identifying
severe, high-fidelity failures.

\section{Conclusion}
\label{sec:conclusion}

\sys is a Pareto-guided search
framework for compositional
robustness testing,
formulated as a bi-objective
problem over performance degradation
$\updelta$ and input fidelity $\uprho$.
Rather than maximising failures alone,
\sys targets configurations where
models fail on inputs that still
resemble the originals, highlighting 
candidate robustness
weaknesses.

The benefit of Pareto-guided search
depends on the geometry of the
Pareto front.
Across vision, natural language, and
code-generation benchmarks, \sys recovers
$\mathcal{P}^*$ more efficiently than
non-Pareto baselines when the front
is broad.
Removing either objective reduces
Pareto-front recovery, and input-level
coverage and prioritisation signals cannot
substitute for bi-objective configuration
search.

The geometry of $\mathcal{P}^*$
depends on the choice of
performance and fidelity metrics, which
are modality- and task-dependent.
Because \sys treats these metrics as
interchangeable components, the framework
generalises to any setting where
performance degradation and input fidelity
can be defined.


\paragraph{Limitations and future work.}
\sys operates over discrete,
predefined perturbation families and
severity levels.
We plan to extend \sys to
continuous perturbation spaces.
Whether extreme Pareto points are useful
depends on the testing goal; future
work should help practitioners target
regions of $\mathcal{P}^*$
and choose suitable search algorithms.

\FloatBarrier

\bibliographystyle{named}
\bibliography{ijcai26}
\clearpage
\clearpage
\appendix 
\newpage
\onecolumn

\section*{Appendix}

The main paper reports recall curves and AUC scores but
cannot show, due to space constraints, the detailed structure
of the ground-truth Pareto fronts $\mathcal{P}^*$ across
benchmarks and metric choices.
This appendix provides that evidence.

Recall that each perturbation configuration is a vector
$\bm{\theta}=(\theta_1,\theta_2,\theta_3,\theta_4)$,
where $\theta_i\in\{0,\ldots,5\}$ gives the severity of
perturbation type $T_i$ and $\theta_i{=}0$ means that
$T_i$ is inactive.
The \emph{perturbation order} of a configuration is
the number of active perturbations,
$|\{i:\theta_i>0\}|$.
We use $\Theta_k$ to denote the subset of the
multi-perturbation space containing configurations with
exactly $k$ active perturbations.
For example, $\bm{\theta}{=}(0,5,0,0)$ is a
$\Theta_1$ configuration: it applies $T_2$ at
severity~5, with $T_1$, $T_3$, and $T_4$ inactive.
In contrast, $\bm{\theta}{=}(1,5,1,1)$ is a
$\Theta_4$ configuration, applying all four perturbation
types at severities $1$, $5$, $1$, and $1$.
Thus, higher-order configurations correspond to
compositions of more perturbation types.

Two questions motivate the appendix:
\begin{inparaenum}[\it (i)]
\item
Do configurations in $\mathcal{P}^*$ include high-fidelity
failures rather than only trivially corrupted inputs?
\item
Does the choice of fidelity metric $\uprho$ matter in
practice?
\end{inparaenum}

Figures~\ref{fig:app:ssim:grid} and~\ref{fig:app:kid:grid}
visualise one fixed Tiny-ImageNet example
from our image-classification benchmark
({\footnotesize \imagenetSymb})
under the Pareto-optimal configurations found using
$(\updelta,\uprho_{\mathrm{SSIM}})$ and
$(\updelta,\uprho_{\mathrm{KID}})$, respectively.
For this benchmark,
$T_1$ is speckle noise,
$T_2$ is glass blur,
$T_3$ is brightness,
and $T_4$ is pixelate.
Each image is labelled with the configuration
$\bm{\theta}$ and its corresponding configuration-level
scores $\updelta(\bm{\theta})$ and $\uprho(\bm{\theta})$.
The examples allow visual inspection of the perturbations,
while $\updelta(\bm{\theta})$ reports the performance
degradation induced by that configuration over the test set.

Table~\ref{tab:app:pareto:summary} summarises the
Pareto-front composition for
{\footnotesize \imagenetSymb}
under the two fidelity metrics.
SSIM and KID yield fronts of the same size
($|\mathcal{P}^*|=27$) but different order distributions:
SSIM places more configurations in $\Theta_1$ and
$\Theta_2$, whereas KID places more in $\Theta_3$.
Thus, the fidelity metric affects which
degradation--fidelity trade-offs are Pareto-optimal.

\begin{figure}[p]
\centering
\setlength{\abovecaptionskip}{2pt}
\setlength{\belowcaptionskip}{0pt}

\begin{subfigure}[t]{0.95\textwidth}
  \centering
  \includegraphics[
    width=\linewidth,
  ]{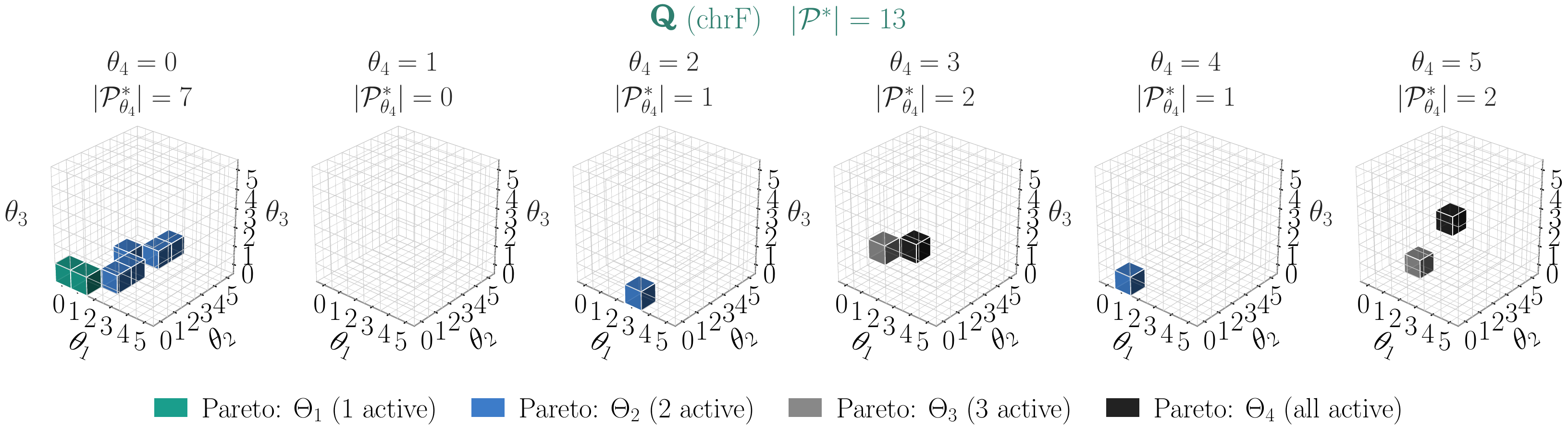}
\end{subfigure}

(a) Paraphrase detection ({\footnotesize \qqpSymb})

\vspace{4pt}

\begin{subfigure}[t]{0.95\textwidth}
  \centering
  \includegraphics[
    width=\linewidth,
  ]{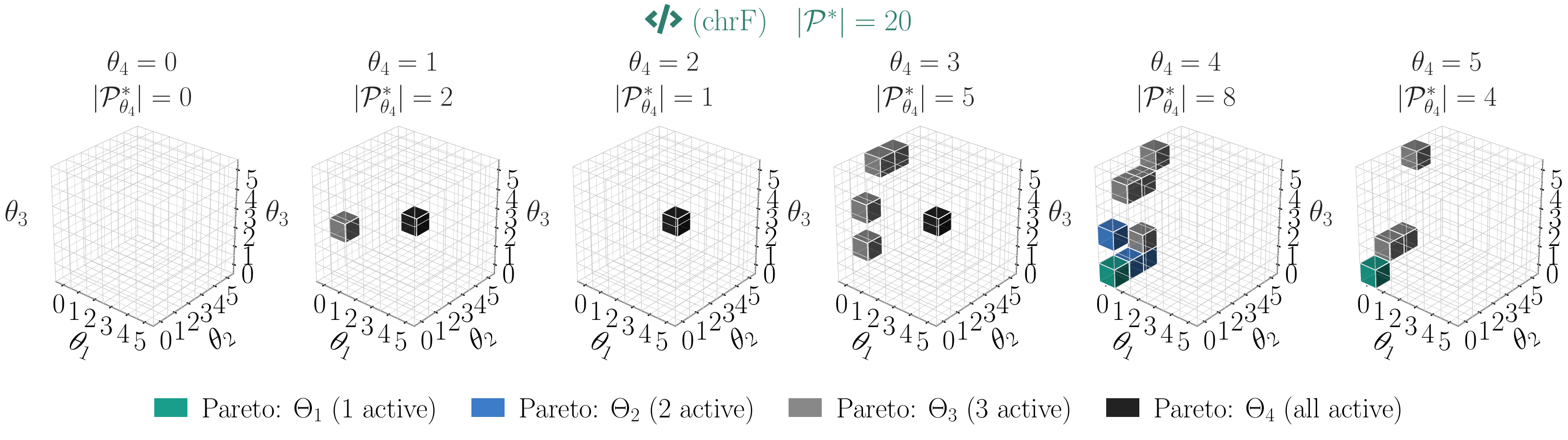}
\end{subfigure}

(b) Code generation (HumanEval) ({\footnotesize \recodeSymb})

\vspace{4pt}

\begin{subfigure}[t]{0.95\textwidth}
  \centering
  \includegraphics[
    width=\linewidth,
  ]{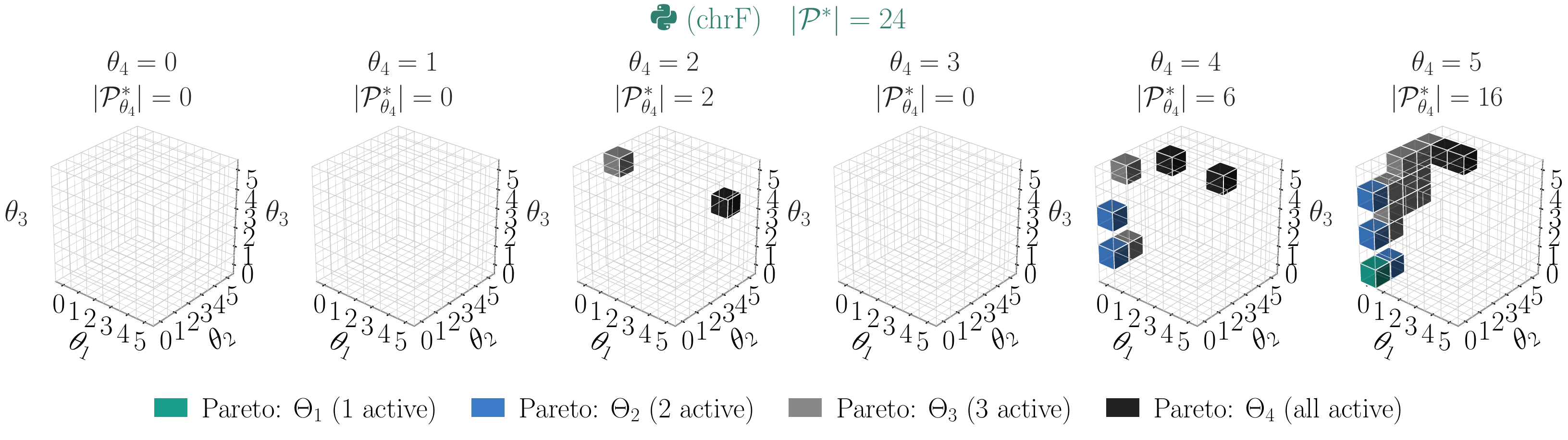}
\end{subfigure}

(c) Code generation (MBPP) ({\footnotesize \mbppSymb})

\vspace{4pt}

\caption{4D voxel visualisations of $\mathcal{P}^*$ for non-vision
benchmarks using chrF as the fidelity metric.}

\label{fig:app:voxel:chrf}
\end{figure}

\begin{figure}[p]
\centering
\setlength{\abovecaptionskip}{2pt}
\setlength{\belowcaptionskip}{0pt}

\begin{subfigure}[t]{0.95\textwidth}
  \centering
  \includegraphics[
    width=\linewidth,
  ]{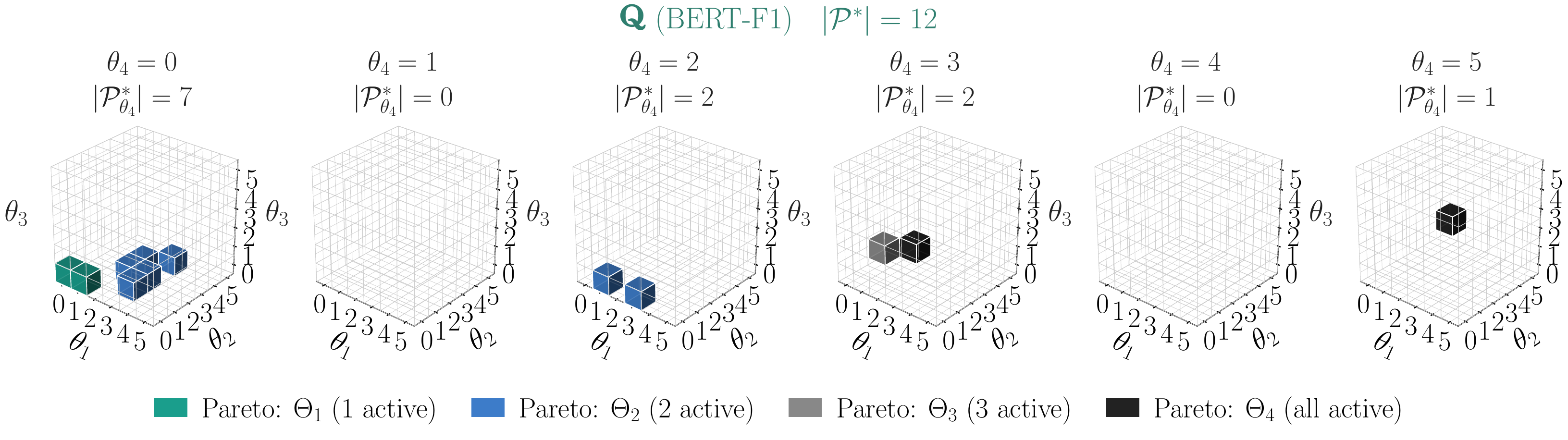}
\end{subfigure}

(a) Paraphrase detection ({\footnotesize \qqpSymb})

\vspace{4pt}

\begin{subfigure}[t]{0.95\textwidth}
  \centering
  \includegraphics[
    width=\linewidth,
  ]{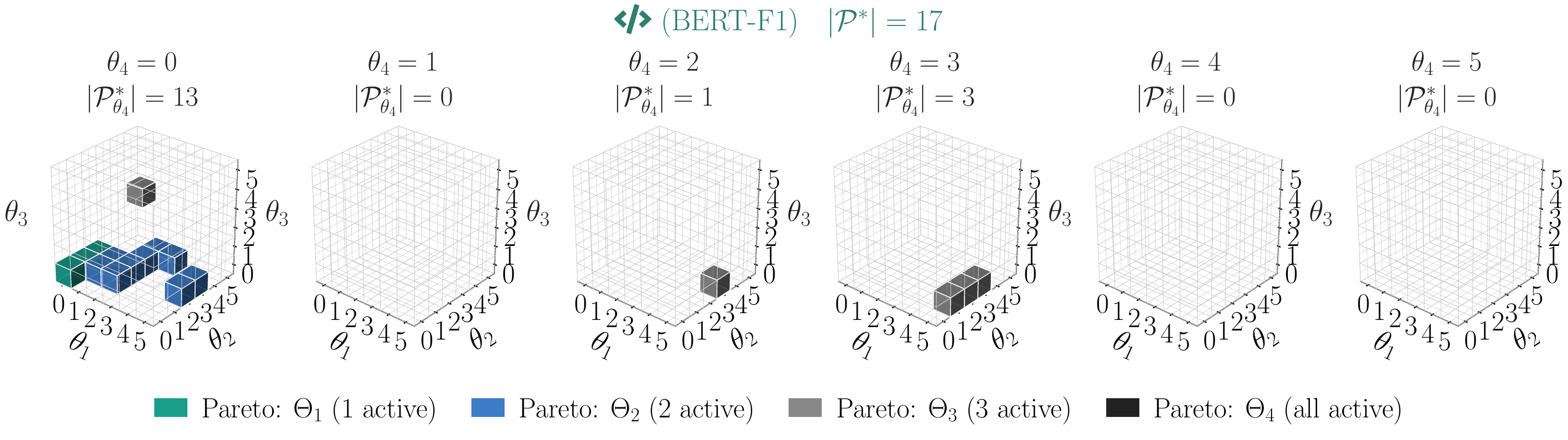}
\end{subfigure}

(b) Code generation (HumanEval) ({\footnotesize \recodeSymb})

\vspace{4pt}

\begin{subfigure}[t]{0.95\textwidth}
  \centering
  \includegraphics[
    width=\linewidth,
  ]{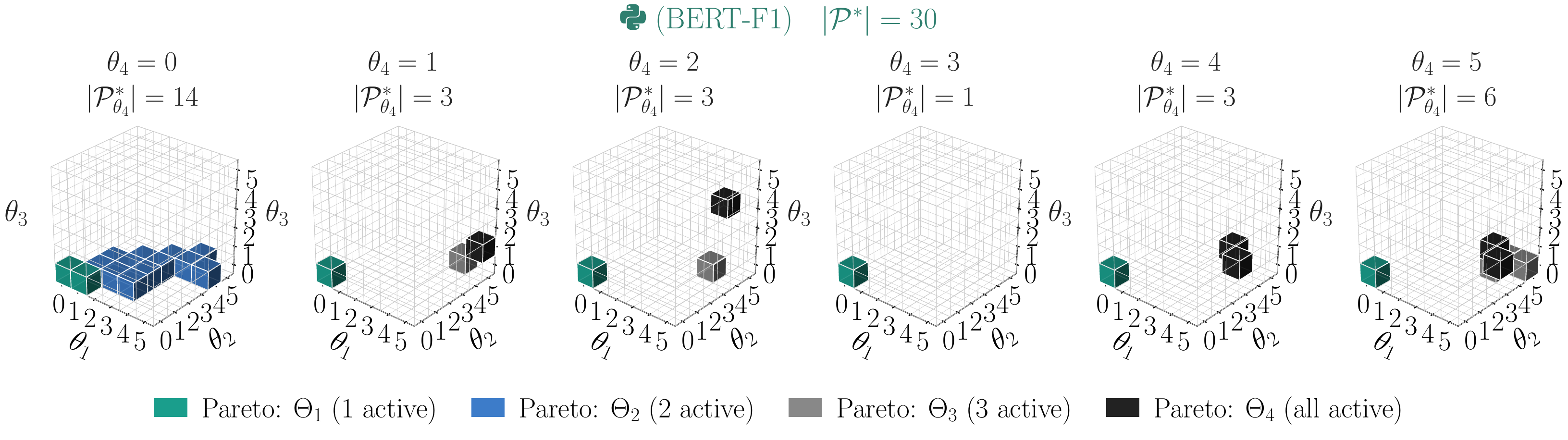}
\end{subfigure}

(c) Code generation (MBPP) ({\footnotesize \mbppSymb})

\vspace{4pt}

\caption{4D voxel visualisations of $\mathcal{P}^*$ for non-vision
benchmarks using BERT-F1 as the fidelity metric.}

\label{fig:app:voxel:bert}
\end{figure}

Figures~\ref{fig:app:voxel:chrf}, and~\ref{fig:app:voxel:bert}
address the second question using 4D voxel plots.
Each plot shows where $\mathcal{P}^*$ lies in the full
$6^4{=}1{,}296$-configuration space, with voxels coloured
by perturbation order.
For non-vision benchmarks, the perturbation types
$T_1,\ldots,T_4$ follow Table~\ref{tab:benchmarks};
the same four-dimensional configuration notation is used.
Together, the plots show that Pareto-optimal configurations
include both low-order and higher-order perturbation
combinations across modalities and fidelity metrics.

\begin{table}[H]
\centering
\small
\setlength{\tabcolsep}{8pt}
\begin{tabular}{lccccc}
\toprule
Metric pair & $\mathcal{P}^*$ & $\Theta_1$ & $\Theta_2$ & $\Theta_3$ & $\Theta_4$ \\
\midrule
$(\updelta,\uprho_{\mathrm{SSIM}})$ & 27 & 9 & 12 & 4 & 2 \\
$(\updelta,\uprho_{\mathrm{KID}})$  & 27 & 4 &  7 & 14 & 2 \\
\bottomrule
\end{tabular}
\caption{Pareto-front composition per metric pair for {\footnotesize \imagenetSymb}.}
\label{tab:app:pareto:summary}
\end{table}




\newcommand{\pimg}[1]{%
  \includegraphics[width=0.13\linewidth]{#1}\hspace{1pt}%
}


\newcommand{\timg}[1]{%
  \includegraphics[width=0.125\textwidth]{#1}%
}

\begin{figure}[p]
\centering
\setlength{\tabcolsep}{1.2pt}
\renewcommand{\arraystretch}{1.15}
\begin{tabular}{ccccccc}
\timg{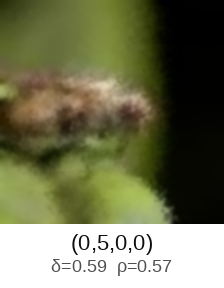}
&
\timg{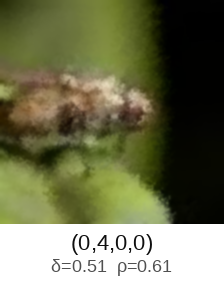}
&
\timg{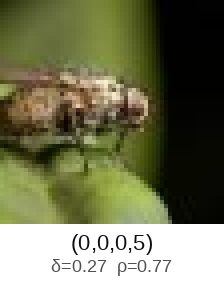}
&
\timg{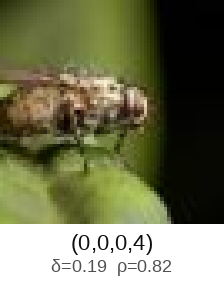}
&
\timg{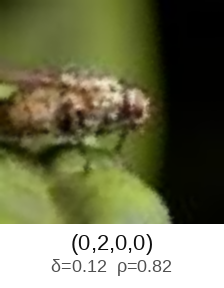}
&
\timg{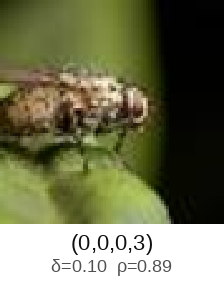}
&
\timg{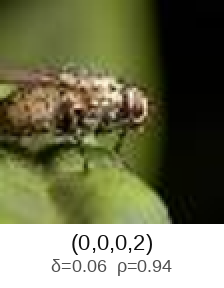}
\\
\timg{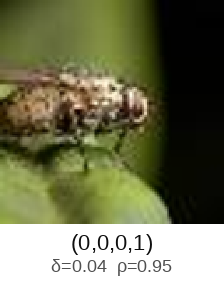}
&
\timg{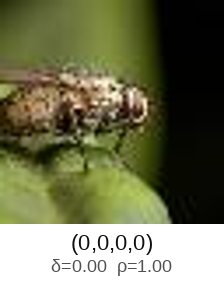}
& & & & & \\
\multicolumn{7}{c}{(a) 9 $\Theta_1$ configurations} 
\vspace{4pt}
\\

\timg{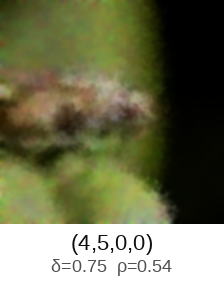}
&
\timg{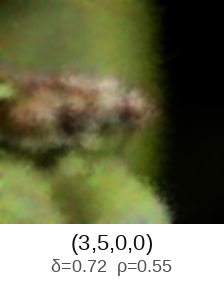}
&
\timg{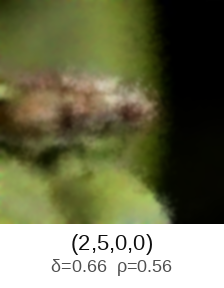}
&
\timg{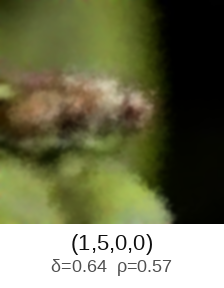}
&
\timg{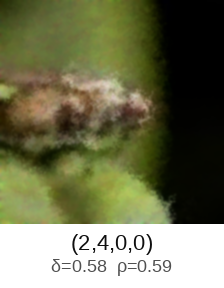}
&
\timg{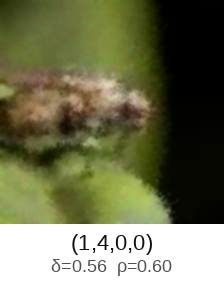}
&
\timg{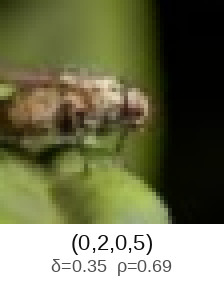}
\\
\timg{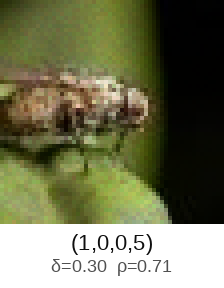}
&
\timg{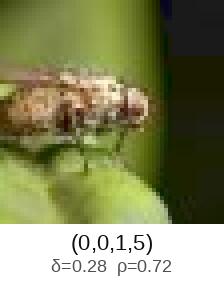}
&
\timg{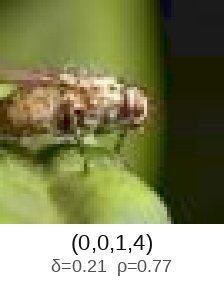}
&
\timg{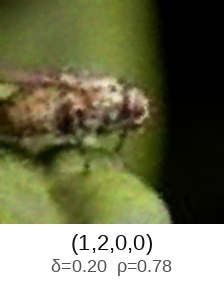}
&
\timg{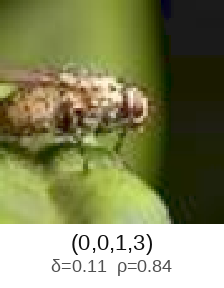}
& & \\
\multicolumn{7}{c}{(b) 12 $\Theta_2$ configurations}
\vspace{4pt}
\\

\timg{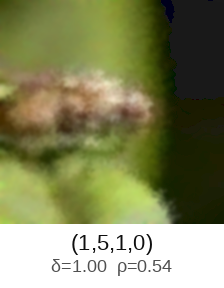}
&
\timg{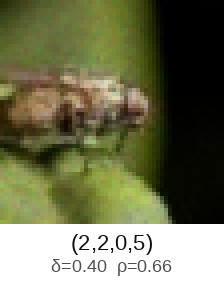}
&
\timg{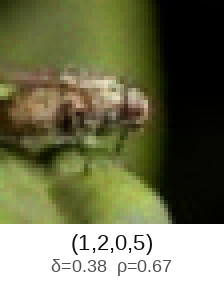}
&
\timg{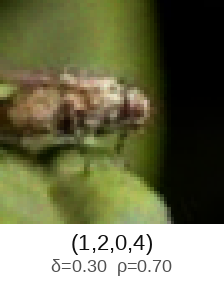}
& & & \\
\multicolumn{7}{c}{(c) 4 $\Theta_3$ configurations}
\vspace{4pt}
\\

\timg{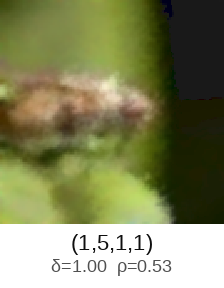}
&
\timg{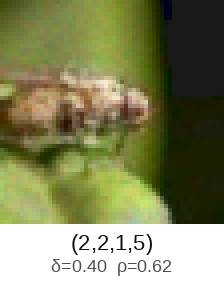}
& & & & & \\
\multicolumn{7}{c}{(d) 2 $\Theta_4$ configurations} \\

\end{tabular}

\caption{$(\updelta,\uprho_{\mathrm{SSIM}})$ Pareto-optimal
configurations for one Tiny-ImageNet example, grouped by active
perturbation count. Each image is labelled with its perturbation
configuration $\bm{\theta}$ and the corresponding
configuration-level scores $\updelta(\bm{\theta})$
and $\uprho(\bm{\theta})$.}
\label{fig:app:ssim:grid}
\end{figure}

\begin{figure}[p]
\centering
\setlength{\tabcolsep}{1.2pt}
\renewcommand{\arraystretch}{1.15}
\begin{tabular}{ccccccc}
\timg{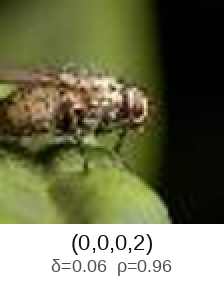}
&
\timg{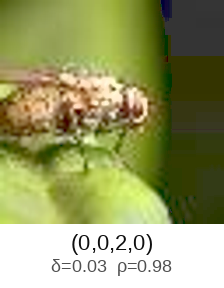}
&
\timg{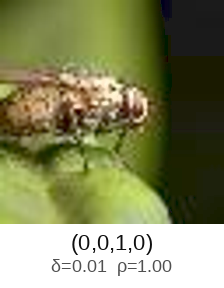}
&
\timg{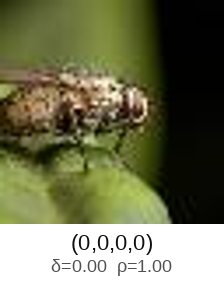}
& & & \\
\multicolumn{7}{c}{(a) 4 $\Theta_1$ configurations}
\vspace{4pt}
\\

\timg{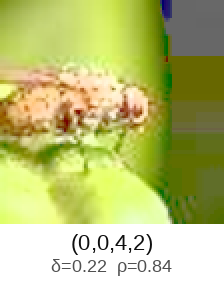}
&
\timg{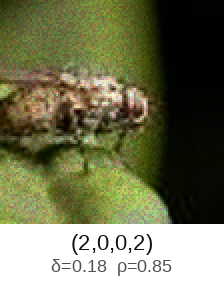}
&
\timg{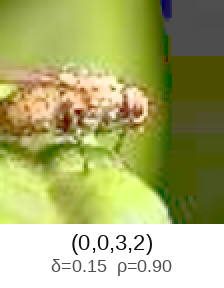}
&
\timg{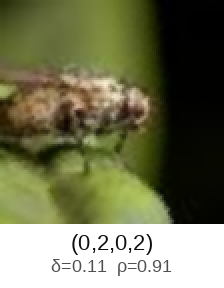}
&
\timg{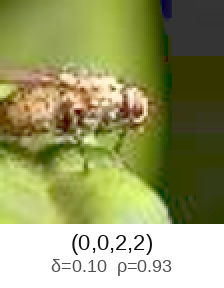}
&
\timg{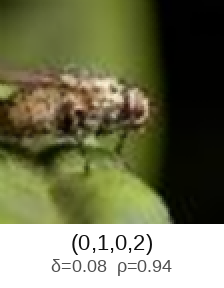}
&
\timg{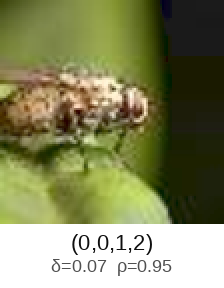}
\\
\multicolumn{7}{c}{(b) 7 $\Theta_2$ configurations}
\vspace{4pt}
\\

\timg{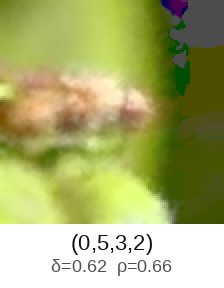}
&
\timg{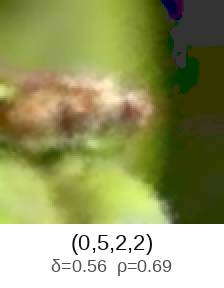}
&
\timg{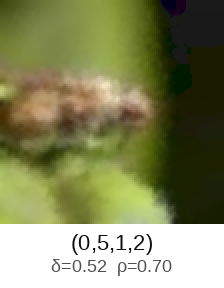}
&
\timg{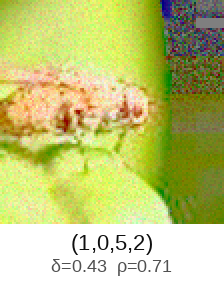}
&
\timg{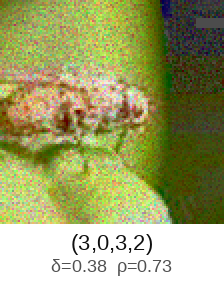}
&
\timg{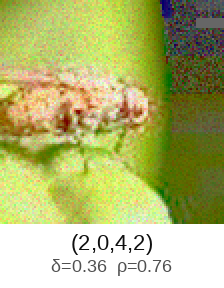}
&
\timg{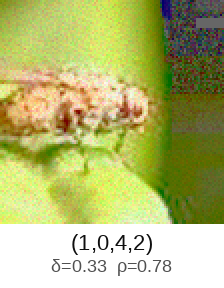}
\\
\timg{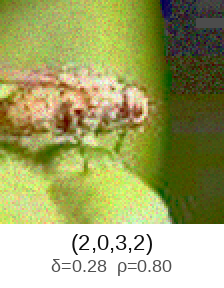}
&
\timg{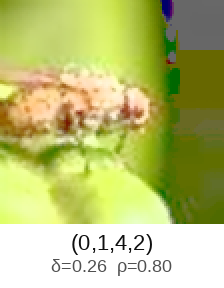}
&
\timg{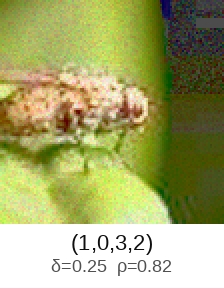}
&
\timg{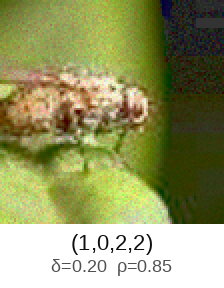}
&
\timg{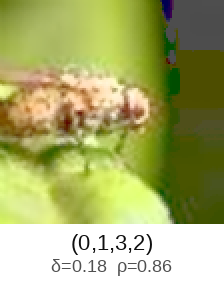}
&
\timg{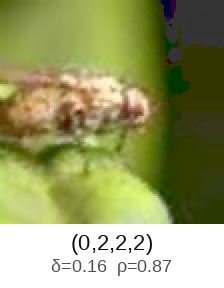}
&
\timg{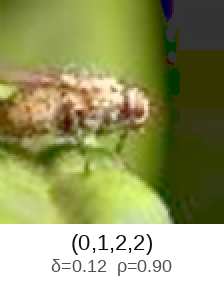}
\\
\multicolumn{7}{c}{(c) 14 $\Theta_3$ configurations}
\vspace{4pt}
\\

\timg{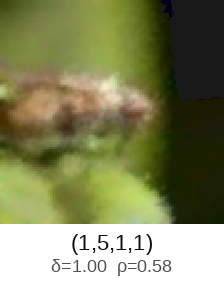}
&
\timg{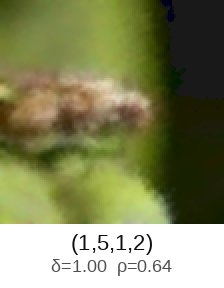}
& & & & & \\
\multicolumn{7}{c}{(d) 2 $\Theta_4$ configurations} \\

\end{tabular}

\caption{$(\updelta,\uprho_{\mathrm{KID}})$ Pareto-optimal
configurations for one Tiny-ImageNet example, grouped by active
perturbation count. Each image is labelled with its perturbation
configuration $\bm{\theta}$ and the corresponding
configuration-level scores $\updelta(\bm{\theta})$
and $\uprho(\bm{\theta})$.}
\label{fig:app:kid:grid}
\end{figure}

\end{document}